\documentclass[11pt,letterpaper]{mystyle}
\usepackage{fancyhdr}

\usepackage[utf8]{inputenc} 
\usepackage[T1]{fontenc}
\usepackage[numbers]{natbib}
\usepackage{adjustbox}
\usepackage{breakurl}    
\usepackage{hyperref}
\usepackage[edges]{forest}
\usepackage{subcaption}
\usepackage{soul}
\usepackage{multirow}
\usepackage[utf8]{inputenc} 
\usepackage{booktabs}       
\usepackage{amsfonts}       
\usepackage{nicefrac}      
\usepackage{microtype}     
\usepackage{xcolor}        
\usepackage{colortbl}
\usepackage{enumitem}
\usepackage{amssymb}
\usepackage{wrapfig}
\usepackage{courier}
\usepackage{bxcoloremoji}
\usepackage{CJKutf8}
\usepackage{booktabs} 
\usepackage{colortbl} 
\usepackage{xcolor}  
\usepackage{graphicx} 
\usepackage{multirow} 
\usepackage{graphicx} 
\usepackage{newfloat}
\usepackage{listings}
\usepackage{algorithm}
\usepackage{algorithmic}
\usepackage{amsmath}
\usepackage{amsfonts}  
\usepackage{enumitem}
\usepackage{array}     
\usepackage{booktabs}  
\usepackage{graphicx}
\usepackage{xcolor}  
\usepackage{caption}
\usepackage{cite} 
\usepackage{colortbl} 
\newcommand{\github}{\raisebox{-1.5pt}{\includegraphics[height=1.05em]{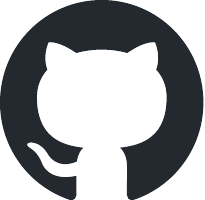}}}

\fancypagestyle{headstyle}{
    \fancyhead[L]{
        \includegraphics[width=80pt]{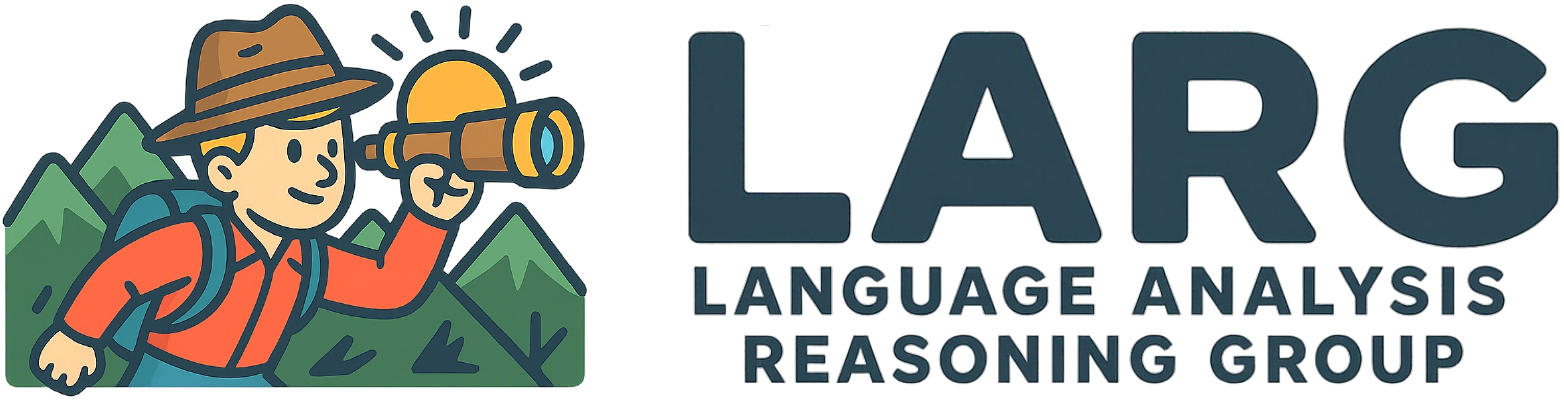}
    }
    \fancyhead[C]{}
    \fancyhead[R]{}

}

\definecolor{hidden-red}{RGB}{205, 44, 36}
\definecolor{hidden-blue}{RGB}{194,232,247}
\definecolor{hidden-orange}{RGB}{243,202,120}
\definecolor{hidden-green}{RGB}{34,139,34}
\definecolor{hidden-pink}{RGB}{255,245,247}
\definecolor{hidden-black}{RGB}{20,68,106}
\definecolor{purple}{RGB}{144,153,196}
\definecolor{yellow}{RGB}{255,228,123}
\definecolor{hidden-yellow}{RGB}{255,248,203}
\definecolor{tkcolor}{RGB}{224,223,255}
\definecolor{darkblue}{rgb}{0, 0.40, 0.75}

\hypersetup{colorlinks=true, citecolor=darkblue, linkcolor=darkblue, urlcolor=darkblue}
\tcbset{
  aibox/.style={
    width=\linewidth,
    top=8pt,
    bottom=4pt,
    colback=blue!6!white,
    colframe=black,
    colbacktitle=black,
    enhanced,
    center,
    attach boxed title to top left={yshift=-0.1in,xshift=0.15in},
    boxed title style={boxrule=0pt,colframe=white,},
  }
}

\definecolor{mypurple}{rgb}{0.878, 0.748, 0.996}
\definecolor{myblue}{rgb}{0.830, 0.839, 0.993}
\definecolor{mygreen}{rgb}{0.821, 0.931, 0.862}

\newtcolorbox{AIbox}[2][]{aibox,title=#2,#1}

\tcbset{
  takeawaybox/.style={
    width=\linewidth,
    top=8pt,
    bottom=4pt,
    colback=hidden-yellow,
    colframe=black,
    colbacktitle=black,
    enhanced,
    center,
    attach boxed title to top left={yshift=-0.1in,xshift=0.15in},
    boxed title style={boxrule=0pt,colframe=white,},
  }
}

\newtcolorbox{TakeawayBox}[2][]{takeawaybox,title=#2,#1}

\title{Aware First, Think Less: Dynamic Boundary Self-Awareness Drives Extreme Reasoning Efficiency in Large Language Models}

\author{
  Qiguang Chen$^{1*}$ \quad Dengyun Peng$^{1*}$ \quad Jinhao Liu$^1$ \quad HuiKang Su$^{1}$ \quad Jiannan Guan$^1$ \quad Libo Qin$^{2, \coloremojicode{2709}}$ \quad Wanxiang Che$^{1, \coloremojicode{2709}}$\\
\normalfont{$^1$ LARG, Research Center for Social Computing and Interactive Robotics, Harbin Institute of Technology,\vspace{-5pt}\\
$^2$ School of Computer Science and Engineering, Central South University\\
}}

\begin{document}

\begin{abstract}
  \vspace{5mm}
  \textbf{\large Abstract:}
  \vspace{2mm}

  Recent advancements in large language models (LLMs) have greatly improved their capabilities on complex reasoning tasks through Long Chain-of-Thought (CoT). However, this  approach often results in substantial redundancy, impairing computational efficiency and causing significant delays in real-time applications. To improve the efficiency, current methods often rely on human-defined difficulty priors, which do not align with the LLM's self-awared difficulty, leading to inefficiencies. In this paper, we introduce the Dynamic Reasoning-Boundary Self-Awareness Framework (DR. SAF), which enables models to dynamically assess and adjust their reasoning depth in response to problem complexity. DR. SAF integrates three key components: Boundary Self-Awareness Alignment, Adaptive Reward Management, and a Boundary Preservation Mechanism. These components allow models to optimize their reasoning processes, balancing efficiency and accuracy without compromising performance. Our experimental results demonstrate that DR. SAF achieves a 49.27\% reduction in total response tokens with minimal loss in accuracy. The framework also delivers a 6.59x gain in token efficiency and a 5x reduction in training time, making it well-suited to resource-limited settings. During extreme training, DR. SAF can even surpass traditional instruction-based models in token efficiency with more than 16\% accuracy improvement.

  \vspace{5mm}

  $^{*}$ \textit{Equal Contribution}
  
  $^{\coloremojicode{2709}}$ \textit{Corresponding Author}

  \vspace{5mm}
  
  \coloremojicode{1F4C5} \textbf{Date}: Aug 01, 2025

  \github{} \textbf{Code Repository}: \href{https://github.com/sfasfaffa/DR_SAF}{https://github.com/sfasfaffa/DR\_SAF}

  \coloremojicode{1F4E7} \textbf{Contact}: \href{mailto:qgchen@ir.hit.edu.cn}{qgchen@ir.hit.edu.cn}, \href{mailto:dypeng@ir.hit.edu.cn}{dypeng@ir.hit.edu.cn}, \href{mailto:car@ir.hit.edu.cn}{car@ir.hit.edu.cn}, \href{mailto:lbqin@csu.edu.cn}{lbqin@csu.edu.cn}

\end{abstract}
\maketitle

\vspace{3mm}
\pagestyle{headstyle}
\thispagestyle{empty}

\begin{figure}[t!]
\centering
\includegraphics[width=0.98\textwidth]{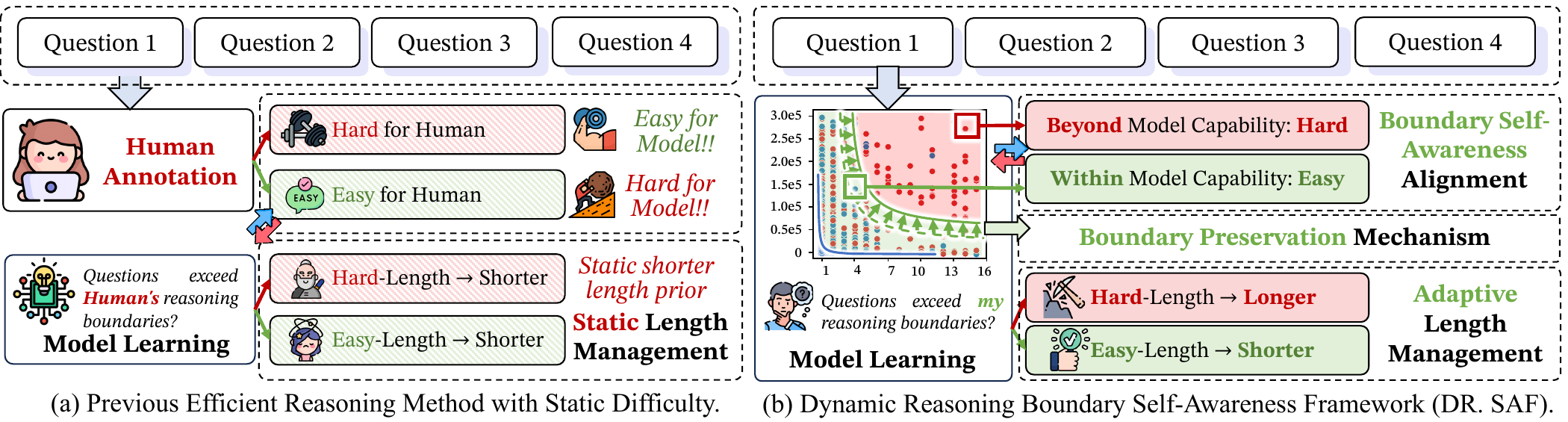} 
\caption{Traditional efficient reasoning training methods (a) primarily determine the difficulty of questions based on human-defined priors, while our dynamic reasoning boundary self-awareness framework (b) judged the difficulty of questions based on model self-awared reasoning boundary.}
\label{fig:intro}
\end{figure}

\vspace{-2mm}\section{Introduction}\vspace{-1mm}
Recent advancements in large language models (LLMs) have demonstrated their remarkable ability to tackle complex reasoning tasks, particularly with the use of Long Chain-of-Thought (Long CoT) techniques~\citep{guo2025deepseek, chen2025towards, li2025system}. In contrast to the short chain-of-thought (Short CoT) typically employed in conventional LLMs~\citep{wei2022chain, qin2023cross,chen2024m3cot}, Long CoT involves a more detailed and progressive process of exploration and reflection based on a given problem. This process is facilitated by inference-time scaling~\citep{guo2025deepseek, zhang2025survey, jaech2024openai}. As a result, Long CoT has significantly advanced areas such as mathematical and logical reasoning. Moreover, it has provided new insights into the role of supervised fine-tuning (SFT) and reinforcement learning (RL) in enhancing the availability of extended reasoning chains~\citep{qin2024o1, min2024imitate}.

While achieving promising performance, such a Long CoT paradigm generates substantial redundant tokens, significantly impairing computational efficiency and leading to unacceptable application latency~\citep{wu2025more,sui2025stop,feng2025efficient}.
To mitigate this issue, several approaches focus on optimizing reasoning length~\citep{tu2025learning,shi2025efficientreinforcementfinetuningadaptive}.
Specifically, a series of studies incorporate compression techniques, such as static pruning thresholds to filter out intermediate tokens~\citep{luo2025o1prunerlengthharmonizingfinetuningo1like} or adaptive routing to more efficient modules~\citep{lu2025prolongedreasoningneedcertaintybased,ling2025fast,liang2025thinkswitcher}.
Other approaches enhance the model's inherent ability to produce concise reasoning paths through techniques~\citep{tu2025learning,shi2025efficientreinforcementfinetuningadaptive}.
For instance, AdaptThink~\citep{zhang2025adaptthinkreasoningmodelslearn} and DAST~\citep{shen2025dast} propose frameworks that dynamically adjust models' reasoning depth based on predefined measures of problem complexity, while \citet{huang2025adactrl} extend these paradigms to human-designed adaptive budgeting.
However, as illustrated in Figure~\ref{fig:intro} (a), current methods often rely on manually designed static priors of difficulty and target length, while neglecting each LLM's intrinsically evolving reasoning boundaries during training~\citep{chen2024unlocking,chen2025rbf++}. As a result, problems initially identified as ``simple'' may remain challenging for an LLM whose capability still needs longer exploration, and those previously judged as ``complex'' might later be handled intuitively by the model by shorter reasoning processes, leading to inefficient reasoning processes and suboptimal performance.

To tackle this challenge, as shown in Figure~\ref{fig:intro} (b), we present the Dynamic Reasoning-Boundary Self-Awareness Framework (DR. SAF), which assesses problem difficulty relative to a model's reasoning capacity. Specifically, DR. SAF consists of three key components: (1) \textbf{Boundary Self-Awareness Alignment} enables LLMs to recognize their real-time reasoning boundaries. This self-awareness allows the model to assess the difficulty of a given question based on its own capabilities, prompting self-guided adjustments in reasoning depth and answer length. (2) \textbf{Adaptive Length Management} further refines efficiency by adapting the reward according to the model's real-time boundaries. It encourages longer exploration beyond the Completely Infeasible Reasoning Boundary (CIRB) and shorter reasoning within the Completely Feasible Reasoning Boundary (CFRB), ensuring that the model does not oversimplify and compromise quality. (3) \textbf{Boundary Preservation Mechanism} maintains stability by preventing the collapse of real-time reasoning boundaries during training, ensuring that all correct responses receive non-negative reinforcement.
These innovations address the traditional trade-off issue between efficiency and accuracy, enabling models to dynamically adjust their reasoning depth based on their capabilities. 

DR. SAF enhances a model's boundary self-awareness, enforces boundary-driven length adaptation, and preserves these boundaries, enabling real-time control of reasoning depth without degrading performance.
When evaluated on six public benchmarks, applying DR. SAF to the distilled Qwen-2.5 model reduces total response tokens by 49.27\% and achieves state-of-the-art token efficiency.
Compared with distilled model, DR. SAF delivers a 6.59x gain in token efficiency. After additional continual training, the extremely compressed DR. SAF model can even surpass traditional instruction-based models in token efficiency and increase accuracy by more than 16\%.
On the distilled Qwen-3 model, DR. SAF reduces training steps by 80\% compared with previous reinforcement-learning-based methods, making it attractive for deployments with limited computational resources.

Our contributions can be summarized as follows:
\begin{itemize}[leftmargin=16pt, itemsep=0pt, topsep=0pt]
    \item We first point out the limitations of existing efficient reasoning methods, which often rely on human-annotated difficulty priors that do not align with LLMs' reasoning requirements. This misalignment leads to inefficient reasoning processes and suboptimal performance.
    \item We propose a novel DR. SAF framework, which enables models to dynamically assess their own reasoning boundaries, adaptively manage length reward signals based on problem feasibility, and prevent models from reasoning boundary collapse.
    \item We systematically demonstrate the effectiveness of DR. SAF through extensive experiments across 6 benchmarks, revealing substantial improvements in efficiency. The extreme speedup can even enable LLMs to surpass the token efficiency of instruction models, while maintaining a 16\% improvement in accuracy.
\end{itemize}

\vspace{-2mm}\section{Preliminaries}\vspace{-1mm}

\subsection{The Efficient Reasoning Objective}\vspace{-1mm}
Given an input $x$, an LLM generates an output $y=\{S, a\}$, which consists of a reasoning step trajectory $S = (s_1, s_2, \dots, s_T)$ and a final answer $a$.

The objective of efficient reasoning is to develop a policy
that minimizes the reasoning path length while preserving accuracy. Formally, the efficient reward is expressed as:
\begin{equation}
R_{\text{Eff}}(y|x) = R_{\text{Acc}}(y|x) + \gamma  R_{\text{Len}}(y|x),
\end{equation}
where $R_{\text{Acc}}(y|x)$ provides a reward of 1 when $y$ is correct, $R_{\text{Len}}(y|x) \propto -\ell_y$ is the length reward, which is negatively correlated with the response length $\ell_y$, and $\gamma$ is a constant hyperparameter.

\begin{figure*}[t]
    \centering
    \includegraphics[width=\textwidth]{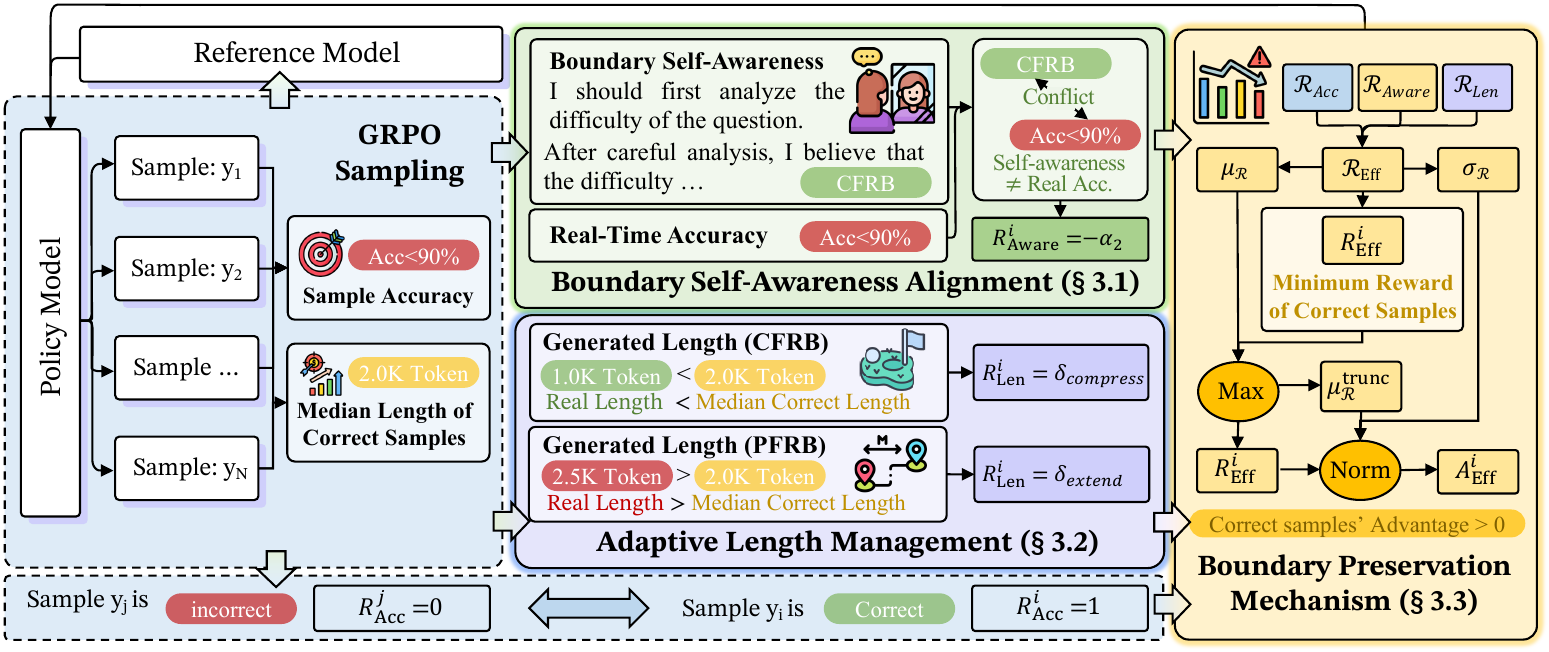} 
    \caption{Main pipeline of Dynamic Reasoning-Boundary Self-Awareness Framework (DR. SAF), including Boundary Self-Awareness Alignment (BSA), Adaptive Length Management (ALM), and Boundary Preservation Mechanism (BPM).}
    \label{fig:framework}
\end{figure*}

\vspace{-2mm}\subsection{Group Relative Policy Optimization}\vspace{-1mm}

We utilize Group Relative Policy Optimization (GRPO) to optimize LLMs, an efficient, critic-free reinforcement learning method that reduces memory and computational costs. Specifically, GRPO operates through group-wise advantage estimation. For a given input, the policy model $\pi_\theta$ generates a group of $k$ outputs, $\mathcal{Y} = \{y_1, \dots, y_k\}$, evaluated by a reward function to yield reward group $\mathcal{R}_{\text{Eff}} = \{R_{\text{Eff}}(y_i|x)\}_{i=1}^k$. The advantage for each output is computed by normalizing its reward against the group’s statistics:
\begin{equation}
\mathcal{A}(\mathcal{Y}|x) = \frac{\mathcal{R}_{\text{Eff}} - \mu_{\mathcal{R}}}{\sigma_{\mathcal{R}} + \epsilon},
\end{equation}
where $\mu_{\mathcal{R}}$ and $\sigma_{\mathcal{R}}$ denote the mean and standard deviation of the rewards group $\mathcal{R}_{\text{Eff}}$, respectively. $\epsilon$ is a small constant introduced to prevent division by zero.
The policy is refined by minimizing the GRPO loss $\mathcal{L}_{\text{GRPO}}$, which amplifies actions with high advantage and penalizes those with low advantage.

\vspace{-2mm}\section{Methodology}\vspace{-1mm}

To compress efficiently without sacrificing accuracy, we introduce a three-module framework (see Figure~\ref{fig:framework}): (1) Boundary Self-Awareness Alignment enables model to gauge question difficulty. (2) Adaptive Length Management applies a discrete length-reward schedule that scales with difficulty and reasoning bounds, preventing harmful over-compression. (3) Boundary Preservation Mechanism stabilizes optimization through advantage reshaping. Together, these modules, collectively called DR. SAF, offer a theory-guided solution to balance efficiency and accuracy. Formal proofs of each module's effectiveness are in the Appendix~\ref{append:proof}.

\vspace{-2mm}\subsection{Boundary Self-Awareness Alignment}\vspace{-1mm}
First, the model develops self-awareness of whether a given task falls within its reasoning capabilities. 
As shown in the green area of Figure \ref{fig:framework}, the model calibrates perceived task difficulty against its real-time accuracy; any gap between expected and observed performance incurs a reward penalty.
Specifically, we employ Boundary Self-Awareness Alignment (BSA) with a format-based reward. Guided by carefully designed prompts, the model evaluates its proficiency on each problem: Inspired by \citet{chen2024unlocking}, if it determines a problem to be fully mastered, it classifies it as within its Completely Feasible Reasoning Boundary (CFRB) and provides more concise solution; otherwise, it assigns the problem to its Partially Feasible Reasoning Boundary (PFRB), initiating a deeper reasoning process.

Next, to enable adaptive boundary awareness, BSA assesses the model's accuracy across multiple runs of the same problem. Following \citet{chen2024unlocking},  we label problems with accuracy above 90\% as CFRB and those below 90\% as PFRB. When the model's boundary classification aligns with the problem's true difficulty, resulting in a correct CFRB or PFRB judgment followed by a correct answer, it receives a positive reward. Conversely, if it mislabels a PFRB problem as CFRB and then answers incorrectly, it incurs a negative reward.

Formally, the reward function is defined as:
\begin{equation}
R_{\text{Aware}}(y|x) = 
\begin{cases} 
+\alpha_1, & \text{if } \text{Acc}(\mathcal{Y}|x) \ge 90\% \land \text{Aware}(x) < \texttt{CFRB};\\
+\alpha_1, & \text{if } \text{Acc}(\mathcal{Y}|x) < 90\% \land \texttt{CFRB}\! \le\! \text{Aware}(x) \!\le\! \texttt{PFRB}; \\
-\alpha_2, & \text{otherwise}, \\
\end{cases}
\end{equation}
where $\alpha_1$ and $\alpha_2$ are positive constants that scale the rewards, $\text{Acc}(\mathcal{Y}|x)$ is the model's empirical accuracy on input $x$, and $\text{Aware}(x)$ denotes the model's self-assessed difficulty for $x$. This framework continuously calibrates the model's self-awareness of reasoning boundaries based on performance feedback.

\vspace{-2mm}\subsection{Adaptive Length Management}\vspace{-1mm}
Unlike traditional compression tasks, which focus on unified length penalties, Adaptive Length Management (ALM) introduces staged incentives to generate suitable response lengths. As illustrated in the purple area of Figure~\ref{fig:framework}, tasks in CFRB that LLM already masters receive compression rewards, driving concise reasoning. For low-accuracy tasks beyond Completely Infeasible Reasoning Boundary (CIRB), we give extension rewards to longer incorrect answers, prompting the model to elaborate for deeper exploration.

Specifically, based on $k$ sampling results, we select a correct sample set $\mathcal{C}$. First, we determine the minimum number of tokens required to maintain a question within the CFRB. Formally, we define $\bar{\ell}_{\texttt{CFRB}}$, the median response length for correct samples in $\mathcal{C}$, as the model's required length for mastery within CFRB.
Based on this, we then define two reward types: (1) the \textbf{\textit{compression reward}} $\delta_{\text{comp}}$  for fully mastered questions. It rewards answers under CFRB whose length $\ell$ should be below the CFRB mean $\bar{\ell}_{\texttt{CFRB}}$. (2) the \textbf{\textit{extension reward}} $\delta_{\text{ext}}$ for exploration-needed questions, those beyond CIRB. Here the answer length $\ell$ should exceed $\bar{\ell}_{\texttt{CFRB}}$ (if no correct sample exists, the length threshold will degrade to the average length of all samples $\bar{\ell}_{\texttt{All}}$). Formally, the adaptive reward for ALM is:
\begin{equation}
R_{\text{Len}}(y|x) = 
\begin{cases} 
\delta_{\text{comp}} & \text{if } \text{Acc}(\mathcal{Y}|x) \!\!>\!\! 90\% \land \ell \!\leq\! \bar{\ell}_{\texttt{CFRB}} \\ 
\delta_{\text{ext}} & \text{if }  \text{Acc}(\mathcal{Y}|x)\!\! <\!\! 10\% \land  \ell\! >\! \bar{\ell}_{\texttt{CFRB}}\\ 
0 & \text{otherwise}
\end{cases}
\end{equation}
where $\delta_{\text{comp}} < R_{\text{Acc}}$ and $\delta_{\text{ext}} < R_{\text{Acc}}$ are positive constants smaller than the accuracy reward $R_{\text{Acc}}$.

\vspace{-2mm}\subsection{Boundary Preservation Mechanism}\vspace{-1mm}

Training models with only length penalties and correctness rewards often leads to a common issue known as boundary collapse. In this case, the model excessively compresses reasoning chains, causing the advantage of correct responses to fall below zero. As a result, valid reasoning boundaries collapse, and infeasible ones arise, both undermining performance and destabilizing training.
As illustrated in yellow part of Figure~\ref{fig:framework}, we address this by enforcing non-negative advantages for all correct responses.

Let \(\mathcal{C}\) represent the set of correct responses, which defines the feasible region. This region includes responses that meet all correctness, length, and awareness criteria within the CFRB, as well as correct responses that may violate secondary preferences (such as length) within the PFRB. Both categories are considered feasible, ensuring that every output \(y_i \in \mathcal{C}\) receives a non-negative advantage.
Formally, for a given input $x$, we sample a group of $k$ outputs, $\mathcal{Y} = \{y_1, \dots, y_k\}$, and calculate the total efficient rewards $\mathcal{R}_{\text{Eff}} = \{R_{\text{Eff}}(y_i|x)\}_{i=1}^k$ as follows:
\begin{equation}
    R_{\text{Eff}}(y_i|x) = R_{\text{Acc}}(y_i|x) + R_{\text{Len}}(y_i|x) + R_{\text{Aware}}(y_i|x).
\end{equation}

The Boundary Preservation Mechanism (BPM) ensures that correct responses, regardless of length, are not unduly suppressed, thereby preventing boundary collapse. Specifically, we utilize the efficient reward function and the method for calculating boundary preservation advantages. To achieve this, truncated-mean normalization is applied to the output sample group. The advantages of boundary preservation mechanism \(A\) are computed as follows:
\begin{equation}
    \mu_{\mathcal{R}}^{\mathrm{trunc}} = \max(\mu_{\mathcal{R}}, \min_{y_i\in \mathcal{C}} R_{\text{Eff}}(y_i|x)), \\
\end{equation}
where $\mathcal{C}$ is the set of correct responses. This step ensures that the decision boundary for correct responses is preserved, preventing the model from assigning negative advantages to correct but length-variant answers.
Next, the boundary preservation advantages are computed as:
\begin{equation}
    \mathcal{A}_{\text{Pre}} (\mathcal{Y}|x) = \frac{\mathcal{R}_{\text{Eff}} - \mu_{\mathcal{R}}^{\mathrm{trunc}}}{\sigma_{\mathcal{R}} + \epsilon},
\end{equation}
where \(\mu_{\mathcal{R}}\) and \(\sigma_{\mathcal{R}}\) are the untruncated mean and standard deviation of reward group $\mathcal{R}_{\text{Eff}}$. By bounding the group mean as $\mu_{\mathcal{R}}^{\mathrm{trunc}}$, we ensure:
\begin{equation}
    \forall y_i \in \mathcal{C}:\quad \mathcal{A}_{\text{Pre}} (y_i|x) = \frac{R_{\text{Eff}(y_i|x)} - \mu_{\mathcal{R}}^{\mathrm{trunc}}}{\sigma_{\mathcal{R}} + \epsilon} \geq 0.
\end{equation}
This safeguard guarantees that correct responses, regardless of length variation, always receive a non-negative advantage. By enforcing this, the Boundary Preservation Mechanism ensures that valid outputs never receive suppressed advantages, thus preventing boundary collapse.

\vspace{-2mm}\section{Experiments}\vspace{-1mm}
\label{sec:experiment}

\subsection{Experimental Setup}\vspace{-1mm}
We utilize verl~\citep{sheng2024hybridflow} as the reinforcement learning framework on 8 A100-80G GPUs. We randomly sample 5,000 instances from the DeepMath103K~\citep{he2025deepmath} as training set,
and trained DR. SAF based on two LLMs, R1-distill-Qwen-2.5-7B~\citep{guo2025deepseek} and R1-distill-Qwen-3-8B~\citep{guo2025deepseek}.  We validate the effectiveness of strategies on AIME24~\citep{aime2024}, GSM8K~\citep{cobbe2021training}, Math-500~\citep{lightman2024lets}, AMC23~\citep{amc2023}, OlympiadBench~\citep{he-etal-2024-olympiadbench}, and AIME25~\citep{aime2025}.
We report three metrics: Accuracy (ACC in \%), average response token length (LEN), and token efficiency (EFF). The Token Efficiency (EFF) is defined as the ratio of Accuracy to Length (EFF = ACC / LEN in \%), serving as an indicator of the correctness and reasoning efficiency trade-off. 

For comprehensive comparison, we adopt three representative paradigms as baselines for efficient reasoning:
(1) \textbf{Prompting Strategies:} Dynasor-CoT~\citep{fu2025reasoning} and DEER~\citep{yang2025dynamic} activate early-exit mechanisms during reasoning; ThinkSwitcher~\citep{liang2025thinkswitcher} trains a switcher to dynamically choose between long and short CoT. (2) 
\textbf{Offline Strategies:} OverThink~\citep{chen2024not} fine-tunes on the shortest generated answers. Spirit~\citep{cui-etal-2025-stepwise} and ConCISE-SimPO~\citep{qiao2025concise} prune tokens based on confidence scores via supervised fine-tuning or direct preference training.
AdaptThink~\citep{zhang2025adaptthinkreasoningmodelslearn} and DAST~\citep{shen2025dast} incorporate human-defined difficulty priors to learn efficient reasoning trajectories.
\textbf{Online Strategies:} Length-Penalty~\citep{arora2025training} encourages compress all outputs; FEDH~\citep{ling2025fast} applies a human-defined length prior to promote concise reasoning process.
Additionally, we compare the token efficiency of instruction models, like Qwen2.5-Ins~\citep{yang2024qwen25} and Qwen2.5-Math~\citep{yang2024qwen25math}.

\begin{table*}[t!]
\centering
\resizebox{\textwidth}{!}{
\begin{tabular}{l | ccc | ccc | ccc | ccc | ccc | ccc}
\toprule
\textbf{Model Name} & \multicolumn{3}{c|}{\textbf{GSM8K}} & \multicolumn{3}{c|}{\textbf{MATH500}} & \multicolumn{3}{c|}{\textbf{AIME24}} & \multicolumn{3}{c|}{\textbf{AMC}} & \multicolumn{3}{c|}{\textbf{OlymBench}} & \multicolumn{3}{c}{\textbf{AIME25}} \\
\cmidrule(lr){2-19}
 & ACC & LEN & EFF & ACC & LEN & EFF & ACC & LEN & EFF & ACC & LEN & EFF & ACC & LEN & EFF & ACC & LEN & EFF \\
\midrule
Qwen2.5-7B-Ins & 90.9 & 279 & 32.58 & 74.2 & 567 & 13.09 & 12.0 & 1016 & 1.18 & 47.5 & 801 & 5.93 & 39.2 & 827 & 4.74 & 7.6 & 1240 & 0.61 \\
Qwen2.5-7B-Math & 93.2 & 439 & 21.23 & 63.4 & 740 & 8.57 & 19.0 & 1429 & 1.33 & 62.5 & 1022 & 6.12 & 31.5 & 1037 & 3.04 & 4.0 & 2562 & 0.16 \\
Qwen2.5-7B-Math-Ins & 95.2 & 323 & 29.47 & 81.4 & 670 & 12.15 & 10.3 & 1363 & 0.76 & 60.0 & 1029 & 5.83 & 38.9 & 1027 & 3.79 & 9.3 & 2087 & 0.45 \\
\midrule
R1-Distill-Qwen2.5-7B & 92.4 & 1833 & 5.04 & 90.8 & 3854 & 2.36 & 49.2 & 10200 & 0.48 & 90.0 & 6476 & 1.39 & 66.1 & 7789 & 0.85 & 35.0 & 10518 & 0.33 \\
\rowcolor[rgb]{0.942, 0.881, 0.998}
\ \ + ThinkSwitcher & \textbf{92.5} & 1389 & 6.66 & 91.3 & 3495 & 2.61 & 48.3 & 7936 & 0.61 & -- & -- & -- & 57.0 & 5147 & 1.11 & 37.5 & 6955 & 0.54 \\
\rowcolor[rgb]{0.942, 0.881, 0.998}
\ \ + Dynasor-CoT & 89.6 & 1285 & 6.97 & 89.4 & 2661 & 3.36 & 46.7 & 12695 & 0.37 & 85.0 & 5980 & 1.42 & -- & -- & -- & -- & -- & -- \\
\rowcolor[rgb]{0.942, 0.881, 0.998} \ \ + DEER & 90.6 & 917 & 9.88 & 89.8 & 2143 & 4.19 & 49.2 & 9839 & 0.50 & 85.0 & 4451 & 1.91 & -- & -- & -- & -- & -- & -- \\
\rowcolor[rgb]{0.928, 0.936, 0.997} \ \ + OverThink & 91.4 & 879 & 10.39 & \textbf{92.9} & 2405 & 3.86 & 50.0 & 9603 & 0.52 & -- & -- & -- & -- & -- & -- & -- & -- & -- \\
\rowcolor[rgb]{0.928, 0.936, 0.997} \ \ + Spirit & 87.2 & 687 & 12.68 & 90.8 & 1765 & 5.14 & 38.3 & 6926 & 0.55 & -- & -- & -- & -- & -- & -- & -- & -- & -- \\
\rowcolor[rgb]{0.928, 0.936, 0.997} \ \ + ConCISE-SimPO & 92.1 & 715 & 12.88 & 91.0 & 1945 & 4.68 & 48.3 & 7745 & 0.62 & -- & -- & -- & -- & -- & -- & -- & -- & -- \\
\rowcolor[rgb]{0.928, 0.936, 0.997} \ \ + DAST & 86.7 & 459 & 18.89 & 89.6 & 2162 & 4.14 & 45.6 & 7578 & 0.60 & -- & -- & -- & -- & -- & -- & -- & -- & -- \\
\rowcolor[rgb]{0.928, 0.936, 0.997} \ \ + AdaptThink & 91.0 & 309 & 29.45 & 92.0 & 1875 & 4.91 & \textbf{55.6} & 8599 & 0.65 & 85.0$^{*}$ & 4265$^{*}$ & 1.99$^{*}$ & 58.4$^{*}$ & 5988$^{*}$ & 0.98$^{*}$ & \textbf{38.3$^{*}$} & 10380$^{*}$ & 0.37$^{*}$ \\
\rowcolor[rgb]{0.919, 0.969, 0.938} \ \ + Length-Penalty & 87.2 & 263 & 33.16 & 89.1 & 2121 & 4.20 & 51.9 & 7464 & 0.70 & 82.5 & 4411 & 1.87 & \textbf{59.8} & 4919 & 1.22 & 33.3 & 8902 & 0.37 \\
\rowcolor[rgb]{0.919, 0.969, 0.938} \ \ + FEDH & 90.1 & 218 & 41.33 & 88.5 & 1306 & 6.50 & 42.3 & 7242 & 0.58 & -- & -- & -- & -- & -- & -- & -- & -- & -- \\
\rowcolor[rgb]{0.919, 0.969, 0.938} \ \ + \textbf{DR. SAF} & 88.1 & \textbf{162} & \textbf{54.38} & 88.3 & \textbf{1061} & \textbf{8.32} & 50.6 & \textbf{6288} & \textbf{0.80} & \textbf{90.0} & \textbf{3096} & \textbf{2.91} & 59.4 & \textbf{3259} & \textbf{1.82} & 38.2 & \textbf{6764} & \textbf{0.56} \\
\midrule
R1-Distill-Qwen3-8B & 94.2 & 2135 & 4.41 & 90.6 & 7051 & 1.28 & \textbf{67.9} & 20155 & 0.34 & 83.5 & 11931 & 0.70 & 60.1 & 12895 & 0.47 & \textbf{62.9} & 20992 & 0.30 \\
\rowcolor[rgb]{0.919, 0.969, 0.938} \ \ + FEDH$^{*}$ &\textbf{94.4}& 2014 & 4.69 & 92.6 & 6761 & 1.37 & 61.3 &13463 & 0.42 & 94.7 & 11928 & 0.79 & 63.5& 12353 & 0.51 & 46.7 & 14730 & 0.32 \\
\rowcolor[rgb]{0.919, 0.969, 0.938} \ \ + Length-Penalty$^{*}$ &93.3& 604 & 15.45 &92.4	 &2581 & 3.58  & 63.7 &12303 & 0.52 & 89.2& 6166 & 1.45 & 68.4&	7383 & 0.93& 54.7 & 12446 & 0.44 \\
\rowcolor[rgb]{0.919, 0.969, 0.938} \ \ + \textbf{DR. SAF} & 92.3 & \textbf{521} & \textbf{17.72} & \textbf{93.3} & \textbf{2168} & \textbf{4.30} & 66.0 & \textbf{9807} & \textbf{0.67} & \textbf{95.6} & \textbf{4003} & \textbf{2.39} & \textbf{71.3} & \textbf{5766} & \textbf{1.24} & 57.9 & \textbf{10692} & \textbf{0.54} \\
\bottomrule
\end{tabular}
}
\caption{Performance comparison across mathematical benchmarks. \textbf{Bold} marks the best baseline score per metric. For each method we report its most token-efficient variant. Here, ``\raisebox{1pt}{\colorbox{mypurple}{ \rule[-0.2ex]{0pt}{1.0ex} }}'': prompting strategies, ``\raisebox{1pt}{\colorbox{myblue}{ \rule[-0.2ex]{0pt}{1.0ex} }}'': offline strategies, ``\raisebox{1pt}{\colorbox{mygreen}{ \rule[-0.2ex]{0pt}{1.0ex} }}'': online strategies. Rows are ordered by token efficiency on GSM8K. ``$*$'' indicates results reproduced in this study.}
\label{tab:main_results}

\end{table*}

\vspace{-2mm}\subsection{Experimental Results}\vspace{-1mm}

\paragraph{Offline methods yield high accuracy but are less token-efficient than online methods.} Offline approaches use gold-standard reasoning trajectories, raising accuracy by more than 6\% on AIME24 (see in Table~\ref{tab:main_results}); however, their longer reasoning chains increase token consumption. In contrast, online methods are more economical, forcing fewer tokens and shorter reasoning paths while maintaining competitive accuracy. Thus, whereas offline methods maximize precision, online methods achieve a superior balance between accuracy and token efficiency.\vspace{-12pt}

\paragraph{DR. SAF performs state-of-the-art token efficiency with minimal accuracy degradation.}
We next report the main results on token efficiency, overall efficiency, and accuracy. As shown in Table~\ref{tab:main_results}, DR. SAF demonstrates superior performance in reasoning length and token efficiency. 
DR. SAF reduces the average response token count by 26.53\% compared with Length Penalty.
The gains are most pronounced on GSM8K with all data within CFRB, where DR. SAF delivers over \textbf{90\%} shorter reasoning and nearly \textbf{10x} higher token efficiency than the distilled backbone.\vspace{-12pt}

\paragraph{DR. SAF markedly increases LLM token efficiency relative to static difficulty-based reasoning.}
As shown in Table~\ref{tab:main_results}, DR. SAF is evaluated against two representative static baselines, AdaptThink and DAST, whose difficulty and length are fixed based on human prior. Because these baselines cannot adapt to variations in complexity relative to the model’s capabilities, they often allocate more tokens than necessary. In contrast, DR. SAF dynamically updates its efficiency during reasoning, reducing the average token count by 34.33\%. On GSM8K in particular, it achieves a token-efficiency rate of 54.38\%, outperforming AdaptThink by more than 20\%. This adaptive mechanism consistently delivers higher token efficiency across all benchmarks.\vspace{-12pt}

\paragraph{DR. SAF shows significant performance improvements on stronger LLMs.}
As shown in Table~\ref{tab:main_results}, compared to R1-Distill-Qwen3-8B, DR. SAF achieves comprehensive token efficiency gains: on GSM8K, token efficiency improves from 4.41 to 17.72 (\textbf{302\%} increase) with minimal accuracy trade-off; on MATH500, both accuracy (90.6\% to 93.3\%) and token efficiency (1.28 to 4.30, \textbf{236\%} improvement) increase; on AMC, accuracy rises from 83.5\% to 95.6\% while token efficiency improves by \textbf{241\%}. These results demonstrate that DR. SAF achieves an excellent balance between computational efficiency and performance.

\subsection{Feature Analysis of DR. SAF}
In this section, we analyze the key feature of DR. SAF by addressing three central questions: (1) Can DR. SAF outperform its original instruction backbone in token efficiency? (2) Does DR. SAF offer improved training speed? (3) Does DR. SAF focus solely on reasoning compression without enhancing overall performance?\vspace{-12pt}

\begin{figure*}[t!]
\centering
\includegraphics[width=0.98\textwidth]{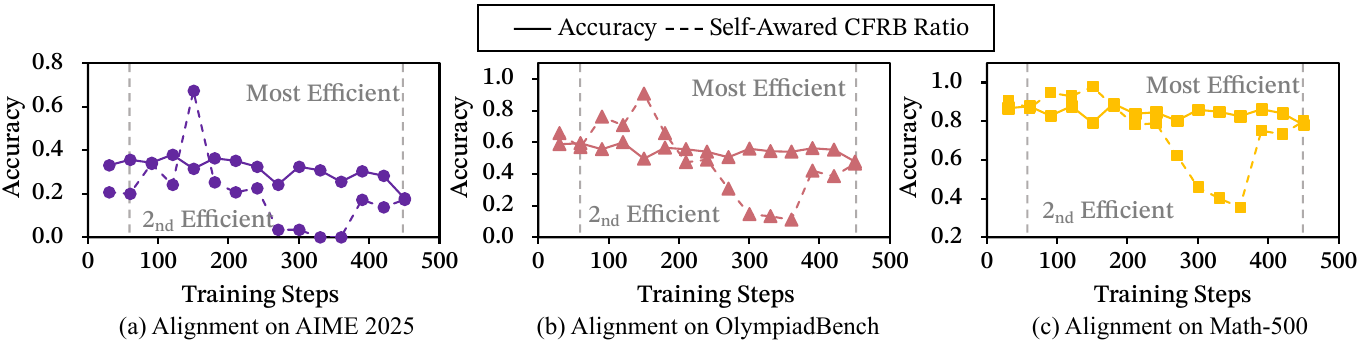} 
    \caption{Training trajectory of BSA, shown as the predicted CFRB ratio plotted against the training steps.}
    \label{fig:boundary-self-awareness}
\end{figure*}

\begin{wrapfigure}{r}{0.50\textwidth}
\centering
    \vspace{-12pt}
    \includegraphics[width=0.48\textwidth]{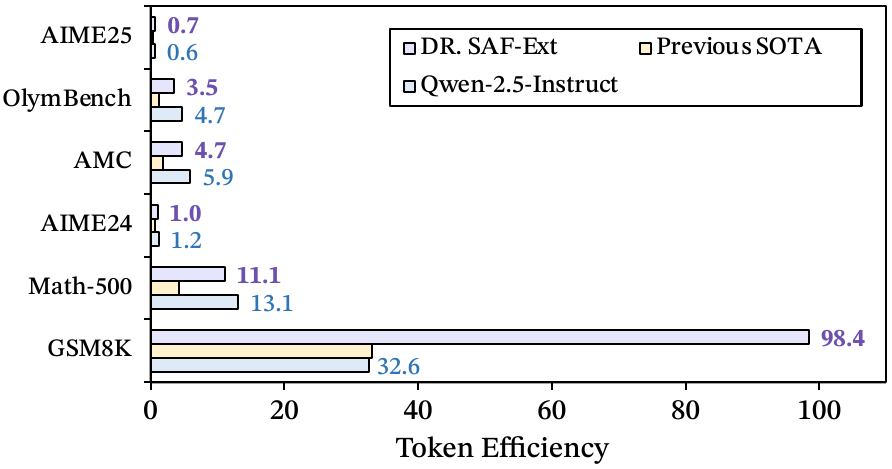} 
    \caption{Comparing the extreme efficiency of DR. SAF (DR. SAF-Ext) with traditional instruction models and current SOTA reasoning efficient techniques.}
    \vspace{-12pt}
    \label{fig:extreme}
\end{wrapfigure}
\paragraph{Answer1: DR. SAF can achieve comparable, or even superior, token efficiency to traditional instruction models across all benchmarks.}
As shown in Figure~\ref{fig:extreme}, earlier techniques match instruction models only on simple datasets such as GSM8K, falling short by over 40\% on more complex tasks. In contrast, the further compressed variant DR. SAF-Ext consistently maintains, and even exceeds, the token efficiency of instruction models across varying task complexities. On the CFRB benchmarks (GSM8K and MATH-500), it doubles the token efficiency of prior methods while matching instruction models. On average, DR. SAF improves token efficiency by 211\% over previous SOTA approaches and increases accuracy by 16.15\% relative to their instruction variance.\vspace{-12pt}

\begin{wrapfigure}{r}{0.52\textwidth}
\centering
    \vspace{-12pt}
    \includegraphics[width=0.48\textwidth]{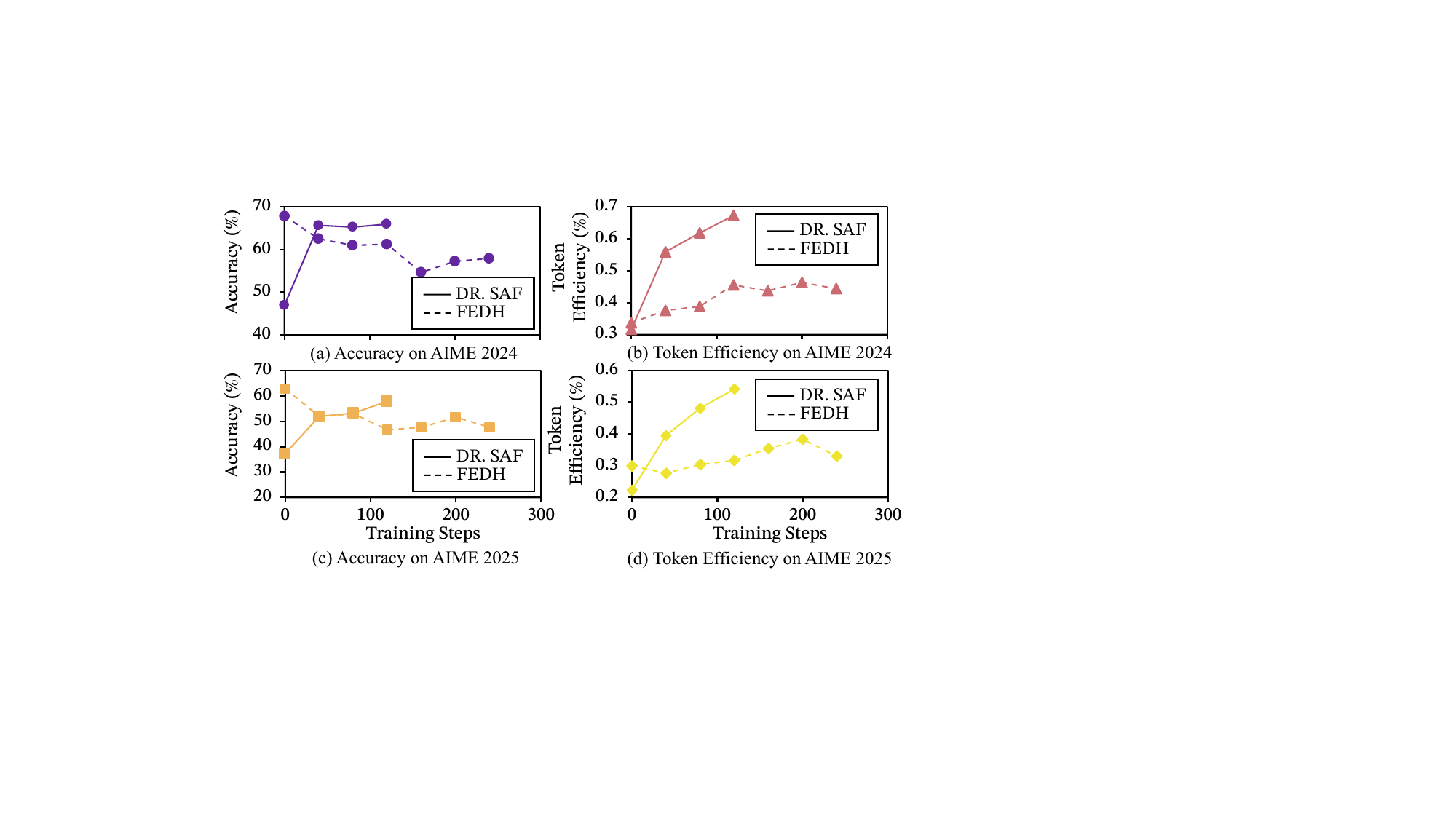} 
    \caption{Training efficiency comparison of DR. SAF vs. FEDH on R1-Distill-Qwen-3-8B.}
    
    \label{fig:speedup}
\end{wrapfigure}
\paragraph{Answer2: DR. SAF achieves significant compression training speedup.}
Compared to previous RL methods, as shown in Figure~\ref{fig:speedup}, DR. SAF reduces training time by up to 5-6 times while maintaining high efficiency. This speedup is particularly evident in large-scale datasets, where DR. SAF minimizes the computational cost associated with model training. Benchmarks indicate that DR. SAF’s compression strategy not only enhances training speed but also ensures minimal loss in model performance, with accuracy even improved. These results demonstrate DR. SAF’s ability to achieve fast and efficient training, making it highly scalable for practical applications.\vspace{-12pt}

\paragraph{Answer3: DR. SAF achieves performance improvement during compression while length penalty-based methods decrease the performance.}
In contrast to length penalty-based methods, which often lead to performance degradation during compression, as shown in Figure~\ref{fig:speedup} (a,c), DR. SAF demonstrates a unique ability to enhance model performance even as it compresses reasoning length. As a result, DR. SAF not only maintains high accuracy but also improves it in many cases.

\subsection{Module Effect Analysis of DR. SAF}
This section evaluates the key effects of each module in DR. SAF with three central questions: (1) Does BSA enhance the model's boundary self-awareness, thereby improving token efficiency?
(2) Does ALM adaptively control token length to simultaneously improve accuracy and efficiency?
(3) Does BPM prevent reasoning boundary collapse during compression training, thus preserving model performance?\vspace{-12pt}

\begin{figure*}[t]
    \centering
    \includegraphics[width=0.98\textwidth]{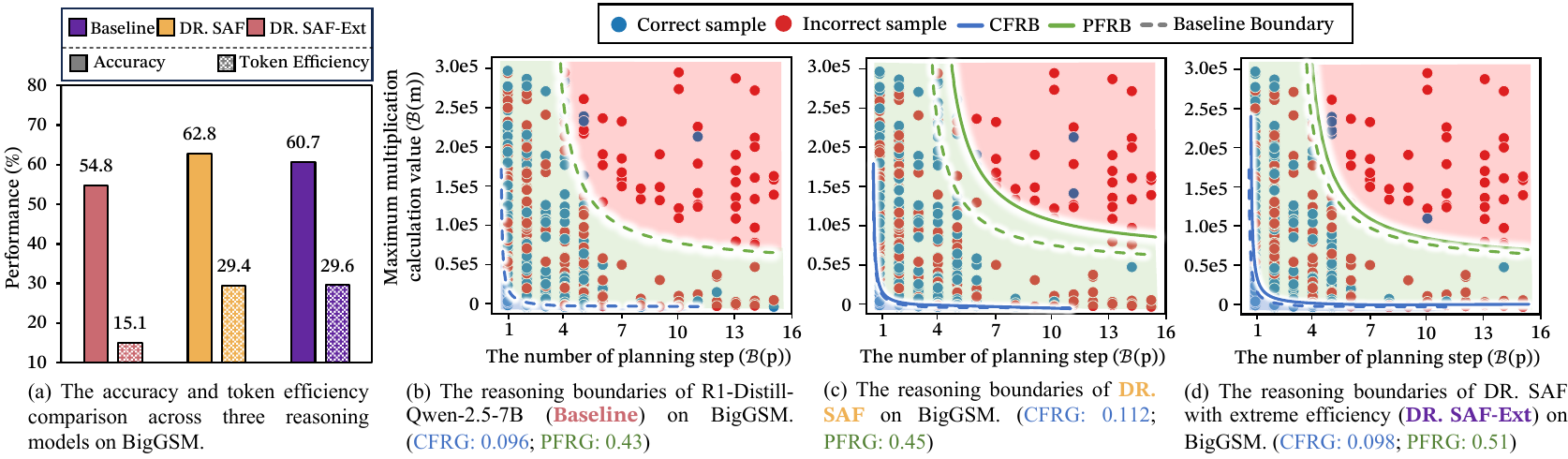} 
    \caption{Boundary Preservation Mechanism's impact on reasoning boundaries on BigGSM~\citep{chen2024unlocking}.}
    \label{fig:bound}
\end{figure*}

\begin{wraptable}{r}{0.4\textwidth}
    \vspace{-12pt}
\centering
\resizebox{0.38\textwidth}{!}{
\begin{tabular}{l | ccc }
    \toprule
    \textbf{Model Name} & ACC$_{\text{AVG}}$ & LEN$_{\text{AVG}}$ & EFF$_{\text{AVG}}$ \\
    \midrule
    \textbf{DR. SAF} & 75.28 & 2773.2 & 13.65 \\
    \midrule
    \ \ w/o BSA & 74.98 & 3219.9 & 8.04 \\
    \ \ w/o ALM & 67.54 & 2105.1 & 11.96 \\
    \ \ w/o BPM & 67.87 & 2543.6 & 13.65 \\
    \bottomrule
    \end{tabular}
    }
    \caption{Ablation analysis of the model's average accuracy, length, and efficiency scores across GSM8K, MATH500, AIME24, AMC, and OlymBench.}
    \vspace{-12pt}
    \label{tab:ablation}
\end{wraptable}

\paragraph{Answer1: Boundary Self-Awareness Alignment is crucial for DR. SAF efficiency.}
We assess the effectiveness of the Boundary Self-Awareness Mechanism by ablating it from the DR. SAF. As shown in Table~\ref{tab:ablation}, token efficiency decreases by more than 40\% without this component. Further, Figure~\ref{fig:boundary-self-awareness} reveals that, during training, the model’s predicted task difficulty progressively aligns with its actual reasoning accuracy. Notably, the models with the highest and second-highest alignment scores also achieve the highest and second-highest levels of token efficiency, respectively.\vspace{-12pt}

\begin{figure*}[t]
    \centering
    \begin{minipage}[b]{0.48\textwidth}
        \centering
        \includegraphics[width=\textwidth]{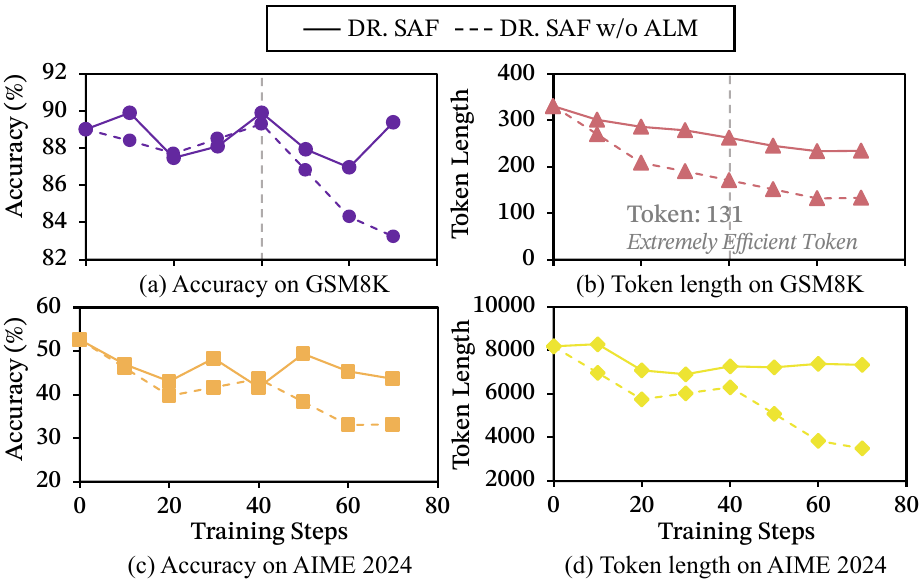} 
    \caption{Trends in accuracy and response length during training with Adaptive Length Management.}
    \label{fig:adaptive-length}
    \end{minipage}
    \hspace{0.05\textwidth}
    \begin{minipage}[b]{0.42\textwidth}
        \centering
        \includegraphics[width=\textwidth]{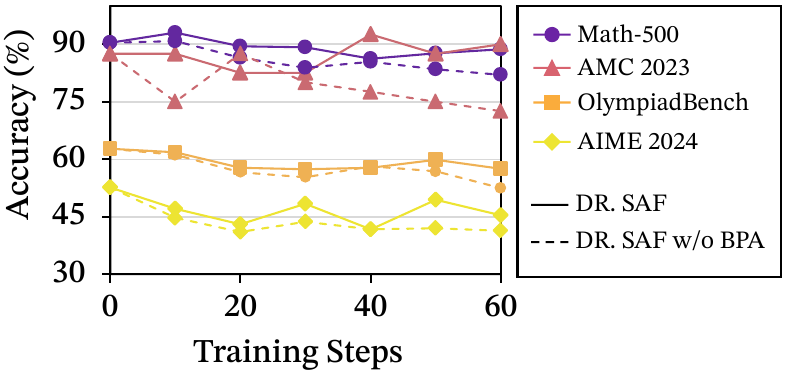} 
        \caption{Accuracy trend produced during training by the boundary-preservation mechanism.}
        \label{fig:bound-change}
    \end{minipage}
\end{figure*}

\paragraph{Answer2: Adaptive Length Management is crucial for both accuracy and efficiency in DR. SAF.}
Replacing Adaptive Length Management with a simple length penalty lowers average response length by 7.74\% and token efficiency score by 1.69 (Table~\ref{tab:ablation}).
Figure~\ref{fig:adaptive-length} further shows that, although the penalty initially shortens reasoning in CFRB (GSM8K), efficiency declines once responses fall below a critical threshold or when tackling harder problems such as AIME. These results confirm that ALM is necessary to sustain the optimal trade-off between brevity and efficiency.\vspace{-12pt}

\paragraph{Answer3: The Boundary Preservation Mechanism effectively mitigates the reasoning boundary collapse of DR. SAF.}
As indicated by the ablation results in Table~\ref{tab:ablation}, removing this module leads to a 7.41\% decrease in accuracy, while token efficiency remains unchanged. This suggests that BPM primarily preserves the model’s reasoning boundaries rather than enhancing token compression. The preservation effect is further supported by the results in Figure~\ref{fig:bound}, where BPM raises Out-of-Domain accuracy on BigGSM from 54.8\% to over 60.7\%. Notably, DR. SAF maintains improved performance even under extreme token compression scenarios (DR. SAF-Ext), highlighting BPM’s robustness across varying compression levels. Consequently, as depicted in Figure~\ref{fig:bound-change}, BPM effectively reduces the extent of boundary collapse in DR. SAF, thereby preventing significant performance degradation across a diverse range of tasks and benchmarks.

\vspace{-2mm}\section{Related work}\vspace{-1mm}
Long Chain-of-Thought (CoT) prompting has substantially advanced domains such as mathematical and logical reasoning~\citep{hu2025owl,hu2025text2world,ji2025mpcc,huang2025mldebugging}. Furthermore, it has offered novel insights into the contributions of supervised fine-tuning (SFT) and reinforcement learning (RL) for improving both the acquisition and exploration of extended reasoning chains~\citep{qin2024o1, min2024imitate,chen2025towards}. Notably, confidence-aware methods let LLMs spend compute only when needed.
During reasoning, techniques such as prolonged reasoning \citep{lu2025prolongedreasoningneedcertaintybased} and dynamic early exit \citep{yang2025dynamic} utilize output probabilities to decide when to stop reasoning~\citep{qiao2025concise}. \citet{eo2025debatenecessaryadaptivemultiagent} uses evaluators for dynamic stopping, and \citep{ding2025dynamicparalleltreesearch} exploits reasoning structure confidence. Additionally, Length-filtered Vote \citep{wu2025more} filters reasoning sequences based on confidence-length correlations.
Other approaches make pre-reasoning decisions. HybridLLM \citep{ding2024hybridllmcostefficientqualityaware}, RouteLLM \citep{ong2025routellmlearningroutellms}, System-1.x \citep{saha2025system1xlearningbalancefast}, and DynaThink \citep{pan-etal-2024-dynathink} train confidence-based triggers to determine reasoning necessity~\citep{luo2025autol2sautolongshortreasoning,zhang2025othinkr1,li2025tldrlongreweightingefficient}. AdaptThink \citep{zhang2025adaptthinkreasoningmodelslearn} optimizes constrained objectives to encourage ``no-thinking'' modes for human-defined easy questions~\citep{huang2025adactrl}.
Early efficiency-oriented RL approaches relied on direct length penalties for model outputs \citep{lou2025adacotparetooptimaladaptivechainofthought,team2025kimi,yeo2025demystifyinglongchainofthoughtreasoning,luo2025o1prunerlengthharmonizingfinetuningo1like,ling2025fast}.
More recent work introduces staged training and adaptive rewards: multi-phase and meta-RL frameworks re-allocate compute and token budgets in real time \citep{tu2025learning,shi2025efficientreinforcementfinetuningadaptive,qu2025optimizingtesttimecomputemeta}.
By linking CoT length to problem difficulty, newer methods keep explanations brief on easy tasks and detailed on hard ones, reducing latency without harming accuracy. Specifically, \citet{an2025don} imposes a human-defined token budget that grows with task complexity, while \citet{ling2025fast} embeds length penalties during training to balance brevity and reasoning depth based on human priors, jointly cutting inference time.

However, traditional efficiency-target methods predominantly focus on optimizing reasoning paths based on fixed or human-defined difficulty levels. In contrast, DR.SAF introduces a self-aware system capable of dynamically adjusting the depth of reasoning according to the model's internal capabilities and the real-time complexity of the task. This approach enhances both efficiency and accuracy.

\section{Conclusion}
In this work, we introduce the Dynamic Reasoning-Boundary Self-Awareness Framework (DR.SAF) to incorporate a self-aware system that adjusts the reasoning depth according to the model’s internal capabilities and real-time task complexity, thereby improving both efficiency and accuracy. Extensive experiments on benchmarks such as Math-500 and AIME demonstrate that DR.SAF cuts response tokens \~50\% and trains 5x faster than long chain-of-thought baselines, yet keeps top-tier accuracy across 6 benchmarks. This framework sets the foundation for more scalable, efficient, and reliable LLMs in real-world applications, balancing reasoning depth with performance.

\bibliographystyle{./refstyle}
\bibliography{ref}

\begin{thebibliography}{59}
\providecommand{\natexlab}[1]{#1}
\providecommand{\url}[1]{\texttt{#1}}
\expandafter\ifx\csname urlstyle\endcsname\relax
  \providecommand{\doi}[1]{doi: #1}\else
  \providecommand{\doi}{doi: \begingroup \urlstyle{rm}\Url}\fi

\bibitem[AIME(2024)]{aime2024}
AIME.
\newblock American invitational mathematics examination (aime) aime 2024-i \& ii, 2024.
\newblock URL \url{https://huggingface.co/datasets/Maxwell-Jia/AIME_2024}.

\bibitem[AIME(2025)]{aime2025}
AIME.
\newblock American invitational mathematics examination (aime) 2025-i \& ii, 2025.
\newblock URL \url{https://huggingface.co/datasets/opencompass/AIME2025}.

\bibitem[AMC(2023)]{amc2023}
AMC.
\newblock American mathematics competitions, 2023.
\newblock URL \url{https://artofproblemsolving.com/wiki/index.php/AMC_Problems_and_Solutions}.

\bibitem[An et~al.(2025)An, Wang, Zhou, and Hsieh]{an2025don}
Sohyun An, Ruochen Wang, Tianyi Zhou, and Cho-Jui Hsieh.
\newblock Don't think longer, think wisely: Optimizing thinking dynamics for large reasoning models.
\newblock \emph{arXiv preprint arXiv:2505.21765}, 2025.

\bibitem[Arora and Zanette(2025)]{arora2025training}
Daman Arora and Andrea Zanette.
\newblock Training language models to reason efficiently.
\newblock \emph{arXiv preprint arXiv:2502.04463}, 2025.

\bibitem[Chen et~al.(2024)Chen, Qin, Wang, Zhou, and Che]{chen2024unlocking}
Qiguang Chen, Libo Qin, Jiaqi Wang, Jingxuan Zhou, and Wanxiang Che.
\newblock Unlocking the capabilities of thought: A reasoning boundary framework to quantify and optimize chain-of-thought.
\newblock \emph{Advances in Neural Information Processing Systems}, 37:\penalty0 54872--54904, 2024.

\bibitem[Chen et~al.(2024{\natexlab{2}})Chen, Qin, Zhang, Chen, Xu, and Che]{chen2024m3cot}
Qiguang Chen, Libo Qin, Jin Zhang, Zhi Chen, Xiao Xu, and Wanxiang Che.
\newblock M3cot: A novel benchmark for multi-domain multi-step multi-modal chain-of-thought.
\newblock In \emph{Proceedings of the 62nd Annual Meeting of the Association for Computational Linguistics (Volume 1: Long Papers)}, pages 8199--8221, 2024{\natexlab{2}}.

\bibitem[Chen et~al.(2025)Chen, Qin, Liu, Liao, Wang, Zhou, and Che]{chen2025rbf++}
Qiguang Chen, Libo Qin, Jinhao Liu, Yue Liao, Jiaqi Wang, Jingxuan Zhou, and Wanxiang Che.
\newblock Rbf++: Quantifying and optimizing reasoning boundaries across measurable and unmeasurable capabilities for chain-of-thought reasoning.
\newblock \emph{arXiv preprint arXiv:2505.13307}, 2025.

\bibitem[Chen et~al.(2025{\natexlab{2}})Chen, Qin, Liu, Peng, Guan, Wang, Hu, Zhou, Gao, and Che]{chen2025towards}
Qiguang Chen, Libo Qin, Jinhao Liu, Dengyun Peng, Jiannan Guan, Peng Wang, Mengkang Hu, Yuhang Zhou, Te~Gao, and Wanxiang Che.
\newblock Towards reasoning era: A survey of long chain-of-thought for reasoning large language models.
\newblock \emph{arXiv preprint arXiv:2503.09567}, 2025{\natexlab{2}}.

\bibitem[Chen et~al.(2025{\natexlab{3}})Chen, Qin, Liu, Peng, Wang, Hu, Chen, Che, and Liu]{chen2025ecm}
Qiguang Chen, Libo Qin, Jinhao Liu, Dengyun Peng, Jiaqi Wang, Mengkang Hu, Zhi Chen, Wanxiang Che, and Ting Liu.
\newblock Ecm: A unified electronic circuit model for explaining the emergence of in-context learning and chain-of-thought in large language model.
\newblock \emph{arXiv preprint arXiv:2502.03325}, 2025{\natexlab{3}}.

\bibitem[Chen et~al.(2024{\natexlab{3}})Chen, Xu, Liang, He, Pang, Yu, Song, Liu, Zhou, Zhang, et~al.]{chen2024not}
Xingyu Chen, Jiahao Xu, Tian Liang, Zhiwei He, Jianhui Pang, Dian Yu, Linfeng Song, Qiuzhi Liu, Mengfei Zhou, Zhuosheng Zhang, et~al.
\newblock Do not think that much for 2+ 3=? on the overthinking of o1-like llms.
\newblock \emph{arXiv preprint arXiv:2412.21187}, 2024{\natexlab{3}}.

\bibitem[Cobbe et~al.(2021)Cobbe, Kosaraju, Bavarian, Chen, Jun, Kaiser, Plappert, Tworek, Hilton, Nakano, et~al.]{cobbe2021training}
Karl Cobbe, Vineet Kosaraju, Mohammad Bavarian, Mark Chen, Heewoo Jun, Lukasz Kaiser, Matthias Plappert, Jerry Tworek, Jacob Hilton, Reiichiro Nakano, et~al.
\newblock Training verifiers to solve math word problems.
\newblock \emph{arXiv preprint arXiv:2110.14168}, 2021.

\bibitem[Cui et~al.(2025)Cui, He, Zeng, Liu, Tang, Dai, Han, Luo, Huang, Li, Wang, Xing, Tang, and He]{cui-etal-2025-stepwise}
Yingqian Cui, Pengfei He, Jingying Zeng, Hui Liu, Xianfeng Tang, Zhenwei Dai, Yan Han, Chen Luo, Jing Huang, Zhen Li, Suhang Wang, Yue Xing, Jiliang Tang, and Qi~He.
\newblock Stepwise perplexity-guided refinement for efficient chain-of-thought reasoning in large language models.
\newblock In Wanxiang Che, Joyce Nabende, Ekaterina Shutova, and Mohammad~Taher Pilehvar, editors, \emph{Findings of the Association for Computational Linguistics: ACL 2025}, pages 18581--18597, Vienna, Austria, July 2025. Association for Computational Linguistics.
\newblock ISBN 979-8-89176-256-5.
\newblock URL \url{https://aclanthology.org/2025.findings-acl.956/}.

\bibitem[Ding et~al.(2024)Ding, Mallick, Wang, Sim, Mukherjee, Ruhle, Lakshmanan, and Awadallah]{ding2024hybridllmcostefficientqualityaware}
Dujian Ding, Ankur Mallick, Chi Wang, Robert Sim, Subhabrata Mukherjee, Victor Ruhle, Laks V.~S. Lakshmanan, and Ahmed~Hassan Awadallah.
\newblock Hybrid llm: Cost-efficient and quality-aware query routing, 2024.
\newblock URL \url{https://arxiv.org/abs/2404.14618}.

\bibitem[Ding et~al.(2025)Ding, Jiang, Liu, Jing, Guo, Wang, Zhang, Wang, Liu, Du, Liu, and Tao]{ding2025dynamicparalleltreesearch}
Yifu Ding, Wentao Jiang, Shunyu Liu, Yongcheng Jing, Jinyang Guo, Yingjie Wang, Jing Zhang, Zengmao Wang, Ziwei Liu, Bo~Du, Xianglong Liu, and Dacheng Tao.
\newblock Dynamic parallel tree search for efficient llm reasoning, 2025.
\newblock URL \url{https://arxiv.org/abs/2502.16235}.

\bibitem[Eo et~al.(2025)Eo, Moon, Zi, Park, and Lim]{eo2025debatenecessaryadaptivemultiagent}
Sugyeong Eo, Hyeonseok Moon, Evelyn~Hayoon Zi, Chanjun Park, and Heuiseok Lim.
\newblock Debate only when necessary: Adaptive multiagent collaboration for efficient llm reasoning, 2025.
\newblock URL \url{https://arxiv.org/abs/2504.05047}.

\bibitem[Feng et~al.(2025)Feng, Fang, Ma, and Wang]{feng2025efficient}
Sicheng Feng, Gongfan Fang, Xinyin Ma, and Xinchao Wang.
\newblock Efficient reasoning models: A survey.
\newblock \emph{arXiv preprint arXiv:2504.10903}, 2025.

\bibitem[Fu et~al.(2025)Fu, Chen, Zhuang, Fu, Stoica, and Zhang]{fu2025reasoning}
Yichao Fu, Junda Chen, Yonghao Zhuang, Zheyu Fu, Ion Stoica, and Hao Zhang.
\newblock Reasoning without self-doubt: More efficient chain-of-thought through certainty probing.
\newblock In \emph{ICLR 2025 Workshop on Foundation Models in the Wild}, 2025.
\newblock URL \url{https://openreview.net/forum?id=wpK4IMJfdX}.

\bibitem[Guo et~al.(2025)Guo, Yang, Zhang, Song, Zhang, Xu, Zhu, Ma, Wang, Bi, et~al.]{guo2025deepseek}
Daya Guo, Dejian Yang, Haowei Zhang, Junxiao Song, Ruoyu Zhang, Runxin Xu, Qihao Zhu, Shirong Ma, Peiyi Wang, Xiao Bi, et~al.
\newblock Deepseek-r1: Incentivizing reasoning capability in llms via reinforcement learning.
\newblock \emph{arXiv preprint arXiv:2501.12948}, 2025.

\bibitem[He et~al.(2024)He, Luo, Bai, Hu, Thai, Shen, Hu, Han, Huang, Zhang, Liu, Qi, Liu, and Sun]{he-etal-2024-olympiadbench}
Chaoqun He, Renjie Luo, Yuzhuo Bai, Shengding Hu, Zhen Thai, Junhao Shen, Jinyi Hu, Xu~Han, Yujie Huang, Yuxiang Zhang, Jie Liu, Lei Qi, Zhiyuan Liu, and Maosong Sun.
\newblock {O}lympiad{B}ench: A challenging benchmark for promoting {AGI} with olympiad-level bilingual multimodal scientific problems.
\newblock In Lun-Wei Ku, Andre Martins, and Vivek Srikumar, editors, \emph{Proceedings of the 62nd Annual Meeting of the Association for Computational Linguistics (Volume 1: Long Papers)}, pages 3828--3850, Bangkok, Thailand, August 2024. Association for Computational Linguistics.
\newblock \doi{10.18653/v1/2024.acl-long.211}.
\newblock URL \url{https://aclanthology.org/2024.acl-long.211/}.

\bibitem[He et~al.(2025)He, Liang, Xu, Liu, Chen, Wang, Song, Yu, Liang, Wang, et~al.]{he2025deepmath}
Zhiwei He, Tian Liang, Jiahao Xu, Qiuzhi Liu, Xingyu Chen, Yue Wang, Linfeng Song, Dian Yu, Zhenwen Liang, Wenxuan Wang, et~al.
\newblock Deepmath-103k: A large-scale, challenging, decontaminated, and verifiable mathematical dataset for advancing reasoning.
\newblock \emph{arXiv preprint arXiv:2504.11456}, 2025.

\bibitem[Hu et~al.(2025)Hu, Chen, Zou, Lei, Chen, Li, Mu, Zhang, Shao, and Luo]{hu2025text2world}
Mengkang Hu, Tianxing Chen, Yude Zou, Yuheng Lei, Qiguang Chen, Ming Li, Yao Mu, Hongyuan Zhang, Wenqi Shao, and Ping Luo.
\newblock Text2world: Benchmarking large language models for symbolic world model generation.
\newblock \emph{arXiv preprint arXiv:2502.13092}, 2025.

\bibitem[Hu et~al.(2025{\natexlab{2}})Hu, Zhou, Fan, Nie, Xia, Sun, Ye, Jin, Li, Chen, et~al.]{hu2025owl}
Mengkang Hu, Yuhang Zhou, Wendong Fan, Yuzhou Nie, Bowei Xia, Tao Sun, Ziyu Ye, Zhaoxuan Jin, Yingru Li, Qiguang Chen, et~al.
\newblock Owl: Optimized workforce learning for general multi-agent assistance in real-world task automation.
\newblock \emph{arXiv preprint arXiv:2505.23885}, 2025{\natexlab{2}}.

\bibitem[Huang et~al.(2025)Huang, Feng, Chen, Zhao, Cheng, Bai, Zhou, Li, and Qin]{huang2025mldebugging}
Jinyang Huang, Xiachong Feng, Qiguang Chen, Hanjie Zhao, Zihui Cheng, Jiesong Bai, Jingxuan Zhou, Min Li, and Libo Qin.
\newblock Mldebugging: Towards benchmarking code debugging across multi-library scenarios.
\newblock \emph{arXiv preprint arXiv:2506.13824}, 2025.

\bibitem[Huang et~al.(2025{\natexlab{2}})Huang, Wang, Zhong, Su, Feng, Cao, and Fung]{huang2025adactrl}
Shijue Huang, Hongru Wang, Wanjun Zhong, Zhaochen Su, Jiazhan Feng, Bowen Cao, and Yi~R Fung.
\newblock Adactrl: Towards adaptive and controllable reasoning via difficulty-aware budgeting.
\newblock \emph{arXiv preprint arXiv:2505.18822}, 2025{\natexlab{2}}.

\bibitem[Jaech et~al.(2024)Jaech, Kalai, Lerer, Richardson, El-Kishky, Low, Helyar, Madry, Beutel, Carney, et~al.]{jaech2024openai}
Aaron Jaech, Adam Kalai, Adam Lerer, Adam Richardson, Ahmed El-Kishky, Aiden Low, Alec Helyar, Aleksander Madry, Alex Beutel, Alex Carney, et~al.
\newblock Openai o1 system card.
\newblock \emph{arXiv preprint arXiv:2412.16720}, 2024.

\bibitem[Ji et~al.(2025)Ji, Chen, Chen, Wu, Qin, and Che]{ji2025mpcc}
Yiyan Ji, Haoran Chen, Qiguang Chen, Chengyue Wu, Libo Qin, and Wanxiang Che.
\newblock Mpcc: A novel benchmark for multimodal planning with complex constraints in multimodal large language models.
\newblock \emph{arXiv preprint arXiv:2507.23382}, 2025.

\bibitem[Li et~al.(2025)Li, Liang, Tang, Ji, Wang, Xu, W, Huang, Deng, Wu, Gong, Guo, Liu, Yin, and Liu]{li2025tldrlongreweightingefficient}
Zhong-Zhi Li, Xiao Liang, Zihao Tang, Lei Ji, Peijie Wang, Haotian Xu, Xing W, Haizhen Huang, Weiwei Deng, Ying~Nian Wu, Yeyun Gong, Zhijiang Guo, Xiao Liu, Fei Yin, and Cheng-Lin Liu.
\newblock Tl;dr: Too long, do re-weighting for efficient llm reasoning compression, 2025.
\newblock URL \url{https://arxiv.org/abs/2506.02678}.

\bibitem[Li et~al.(2025{\natexlab{2}})Li, Zhang, Zhang, Zhang, Liu, Yao, Xu, Zheng, Wang, Chen, et~al.]{li2025system}
Zhong-Zhi Li, Duzhen Zhang, Ming-Liang Zhang, Jiaxin Zhang, Zengyan Liu, Yuxuan Yao, Haotian Xu, Junhao Zheng, Pei-Jie Wang, Xiuyi Chen, et~al.
\newblock From system 1 to system 2: A survey of reasoning large language models.
\newblock \emph{arXiv preprint arXiv:2502.17419}, 2025{\natexlab{2}}.

\bibitem[Liang et~al.(2025)Liang, Zhong, Yang, and Quan]{liang2025thinkswitcher}
Guosheng Liang, Longguang Zhong, Ziyi Yang, and Xiaojun Quan.
\newblock Thinkswitcher: When to think hard, when to think fast.
\newblock \emph{arXiv preprint arXiv:2505.14183}, 2025.

\bibitem[Lightman et~al.(2024)Lightman, Kosaraju, Burda, Edwards, Baker, Lee, Leike, Schulman, Sutskever, and Cobbe]{lightman2024lets}
Hunter Lightman, Vineet Kosaraju, Yuri Burda, Harrison Edwards, Bowen Baker, Teddy Lee, Jan Leike, John Schulman, Ilya Sutskever, and Karl Cobbe.
\newblock Let's verify step by step.
\newblock In \emph{The Twelfth International Conference on Learning Representations}, 2024.
\newblock URL \url{https://openreview.net/forum?id=v8L0pN6EOi}.

\bibitem[Ling et~al.(2025)Ling, Chen, Zhang, Jiao, Guo, and Cheng]{ling2025fast}
Zehui Ling, Deshu Chen, Hongwei Zhang, Yifeng Jiao, Xin Guo, and Yuan Cheng.
\newblock Fast on the easy, deep on the hard: Efficient reasoning via powered length penalty.
\newblock \emph{arXiv preprint arXiv:2506.10446}, 2025.

\bibitem[Lou et~al.(2025)Lou, Sun, Liang, Qu, Shen, Wang, Li, Yang, and Wu]{lou2025adacotparetooptimaladaptivechainofthought}
Chenwei Lou, Zewei Sun, Xinnian Liang, Meng Qu, Wei Shen, Wenqi Wang, Yuntao Li, Qingping Yang, and Shuangzhi Wu.
\newblock Adacot: Pareto-optimal adaptive chain-of-thought triggering via reinforcement learning, 2025.
\newblock URL \url{https://arxiv.org/abs/2505.11896}.

\bibitem[Lu et~al.(2025)Lu, Yu, Xu, Ran, Tang, Wang, Shan, Fu, Feng, Tang, Wang, and Huang]{lu2025prolongedreasoningneedcertaintybased}
Jinghui Lu, Haiyang Yu, Siliang Xu, Shiwei Ran, Guozhi Tang, Siqi Wang, Bin Shan, Teng Fu, Hao Feng, Jingqun Tang, Han Wang, and Can Huang.
\newblock Prolonged reasoning is not all you need: Certainty-based adaptive routing for efficient llm/mllm reasoning, 2025.
\newblock URL \url{https://arxiv.org/abs/2505.15154}.

\bibitem[Luo et~al.(2025)Luo, Chuang, Wang, Le, Zhong, Liu, Yuan, Sui, Braverman, Chaudhary, and Hu]{luo2025autol2sautolongshortreasoning}
Feng Luo, Yu-Neng Chuang, Guanchu Wang, Hoang Anh~Duy Le, Shaochen Zhong, Hongyi Liu, Jiayi Yuan, Yang Sui, Vladimir Braverman, Vipin Chaudhary, and Xia Hu.
\newblock Autol2s: Auto long-short reasoning for efficient large language models, 2025.
\newblock URL \url{https://arxiv.org/abs/2505.22662}.

\bibitem[Luo et~al.(2025{\natexlab{2}})Luo, Shen, He, Wang, Liu, Li, Tan, Cao, and Tao]{luo2025o1prunerlengthharmonizingfinetuningo1like}
Haotian Luo, Li~Shen, Haiying He, Yibo Wang, Shiwei Liu, Wei Li, Naiqiang Tan, Xiaochun Cao, and Dacheng Tao.
\newblock O1-pruner: Length-harmonizing fine-tuning for o1-like reasoning pruning, 2025{\natexlab{2}}.
\newblock URL \url{https://arxiv.org/abs/2501.12570}.

\bibitem[Min et~al.(2024)Min, Chen, Jiang, Chen, Deng, Hu, Tang, Wang, Cheng, Song, et~al.]{min2024imitate}
Yingqian Min, Zhipeng Chen, Jinhao Jiang, Jie Chen, Jia Deng, Yiwen Hu, Yiru Tang, Jiapeng Wang, Xiaoxue Cheng, Huatong Song, et~al.
\newblock Imitate, explore, and self-improve: A reproduction report on slow-thinking reasoning systems.
\newblock \emph{arXiv preprint arXiv:2412.09413}, 2024.

\bibitem[Ong et~al.(2025)Ong, Almahairi, Wu, Chiang, Wu, Gonzalez, Kadous, and Stoica]{ong2025routellmlearningroutellms}
Isaac Ong, Amjad Almahairi, Vincent Wu, Wei-Lin Chiang, Tianhao Wu, Joseph~E. Gonzalez, M~Waleed Kadous, and Ion Stoica.
\newblock Routellm: Learning to route llms with preference data, 2025.
\newblock URL \url{https://arxiv.org/abs/2406.18665}.

\bibitem[Pan et~al.(2024)Pan, Zhang, Zhang, Liu, Wang, and Li]{pan-etal-2024-dynathink}
Jiabao Pan, Yan Zhang, Chen Zhang, Zuozhu Liu, Hongwei Wang, and Haizhou Li.
\newblock {D}yna{T}hink: Fast or slow? a dynamic decision-making framework for large language models.
\newblock In Yaser Al-Onaizan, Mohit Bansal, and Yun-Nung Chen, editors, \emph{Proceedings of the 2024 Conference on Empirical Methods in Natural Language Processing}, pages 14686--14695, Miami, Florida, USA, November 2024. Association for Computational Linguistics.
\newblock \doi{10.18653/v1/2024.emnlp-main.814}.
\newblock URL \url{https://aclanthology.org/2024.emnlp-main.814/}.

\bibitem[Qiao et~al.(2025)Qiao, Deng, Zeng, Wang, Wei, Meng, Zhou, Ren, and Zhang]{qiao2025concise}
Ziqing Qiao, Yongheng Deng, Jiali Zeng, Dong Wang, Lai Wei, Fandong Meng, Jie Zhou, Ju~Ren, and Yaoxue Zhang.
\newblock Concise: Confidence-guided compression in step-by-step efficient reasoning.
\newblock \emph{arXiv preprint arXiv:2505.04881}, 2025.

\bibitem[Qin et~al.(2023)Qin, Chen, Wei, Huang, and Che]{qin2023cross}
Libo Qin, Qiguang Chen, Fuxuan Wei, Shijue Huang, and Wanxiang Che.
\newblock Cross-lingual prompting: Improving zero-shot chain-of-thought reasoning across languages.
\newblock In \emph{Proceedings of the 2023 Conference on Empirical Methods in Natural Language Processing}, pages 2695--2709, 2023.

\bibitem[Qin et~al.(2024)Qin, Li, Zou, Liu, Xia, Huang, Ye, Yuan, Liu, Li, et~al.]{qin2024o1}
Yiwei Qin, Xuefeng Li, Haoyang Zou, Yixiu Liu, Shijie Xia, Zhen Huang, Yixin Ye, Weizhe Yuan, Hector Liu, Yuanzhi Li, et~al.
\newblock O1 replication journey: A strategic progress report--part 1.
\newblock \emph{arXiv preprint arXiv:2410.18982}, 2024.

\bibitem[Qu et~al.(2025)Qu, Yang, Setlur, Tunstall, Beeching, Salakhutdinov, and Kumar]{qu2025optimizingtesttimecomputemeta}
Yuxiao Qu, Matthew Y.~R. Yang, Amrith Setlur, Lewis Tunstall, Edward~Emanuel Beeching, Ruslan Salakhutdinov, and Aviral Kumar.
\newblock Optimizing test-time compute via meta reinforcement fine-tuning, 2025.
\newblock URL \url{https://arxiv.org/abs/2503.07572}.

\bibitem[Saha et~al.(2025)Saha, Prasad, Chen, Hase, Stengel-Eskin, and Bansal]{saha2025system1xlearningbalancefast}
Swarnadeep Saha, Archiki Prasad, Justin Chih-Yao Chen, Peter Hase, Elias Stengel-Eskin, and Mohit Bansal.
\newblock System-1.x: Learning to balance fast and slow planning with language models, 2025.
\newblock URL \url{https://arxiv.org/abs/2407.14414}.

\bibitem[Shen et~al.(2025)Shen, Zhang, Huang, Shi, Zhang, Yan, Wang, Wang, Liu, and Lian]{shen2025dast}
Yi~Shen, Jian Zhang, Jieyun Huang, Shuming Shi, Wenjing Zhang, Jiangze Yan, Ning Wang, Kai Wang, Zhaoxiang Liu, and Shiguo Lian.
\newblock Dast: Difficulty-adaptive slow-thinking for large reasoning models.
\newblock \emph{arXiv preprint arXiv:2503.04472}, 2025.

\bibitem[Sheng et~al.(2024)Sheng, Zhang, Ye, Wu, Zhang, Zhang, Peng, Lin, and Wu]{sheng2024hybridflow}
Guangming Sheng, Chi Zhang, Zilingfeng Ye, Xibin Wu, Wang Zhang, Ru~Zhang, Yanghua Peng, Haibin Lin, and Chuan Wu.
\newblock Hybridflow: A flexible and efficient rlhf framework.
\newblock \emph{arXiv preprint arXiv: 2409.19256}, 2024.

\bibitem[Shi et~al.(2025)Shi, Wu, Song, Zhou, and Zhao]{shi2025efficientreinforcementfinetuningadaptive}
Taiwei Shi, Yiyang Wu, Linxin Song, Tianyi Zhou, and Jieyu Zhao.
\newblock Efficient reinforcement finetuning via adaptive curriculum learning, 2025.
\newblock URL \url{https://arxiv.org/abs/2504.05520}.

\bibitem[Sui et~al.(2025)Sui, Chuang, Wang, Zhang, Zhang, Yuan, Liu, Wen, Zhong, Chen, et~al.]{sui2025stop}
Yang Sui, Yu-Neng Chuang, Guanchu Wang, Jiamu Zhang, Tianyi Zhang, Jiayi Yuan, Hongyi Liu, Andrew Wen, Shaochen Zhong, Hanjie Chen, et~al.
\newblock Stop overthinking: A survey on efficient reasoning for large language models.
\newblock \emph{arXiv preprint arXiv:2503.16419}, 2025.

\bibitem[Team et~al.(2025)Team, Du, Gao, Xing, Jiang, Chen, Li, Xiao, Du, Liao, et~al.]{team2025kimi}
Kimi Team, Angang Du, Bofei Gao, Bowei Xing, Changjiu Jiang, Cheng Chen, Cheng Li, Chenjun Xiao, Chenzhuang Du, Chonghua Liao, et~al.
\newblock Kimi k1.5: Scaling reinforcement learning with llms.
\newblock \emph{arXiv preprint arXiv:2501.12599}, 2025.

\bibitem[Tu et~al.(2025)Tu, Lin, Zhang, Tian, Li, Lan, and Zhao]{tu2025learning}
Songjun Tu, Jiahao Lin, Qichao Zhang, Xiangyu Tian, Linjing Li, Xiangyuan Lan, and Dongbin Zhao.
\newblock Learning when to think: Shaping adaptive reasoning in r1-style models via multi-stage rl.
\newblock \emph{arXiv preprint arXiv:2505.10832}, 2025.

\bibitem[Wei et~al.(2022)Wei, Wang, Schuurmans, Bosma, Xia, Chi, Le, Zhou, et~al.]{wei2022chain}
Jason Wei, Xuezhi Wang, Dale Schuurmans, Maarten Bosma, Fei Xia, Ed~Chi, Quoc~V Le, Denny Zhou, et~al.
\newblock Chain-of-thought prompting elicits reasoning in large language models.
\newblock \emph{Advances in neural information processing systems}, 35:\penalty0 24824--24837, 2022.

\bibitem[Wu et~al.(2025)Wu, Wang, Du, Jegelka, and Wang]{wu2025more}
Yuyang Wu, Yifei Wang, Tianqi Du, Stefanie Jegelka, and Yisen Wang.
\newblock When more is less: Understanding chain-of-thought length in {LLM}s.
\newblock 2025.
\newblock URL \url{https://openreview.net/forum?id=W8dxn7hBkO}.

\bibitem[Yang et~al.(2024)Yang, Yang, Zhang, Hui, Zheng, Yu, Li, Liu, Huang, Wei, et~al.]{yang2024qwen25}
An~Yang, Baosong Yang, Beichen Zhang, Binyuan Hui, Bo~Zheng, Bowen Yu, Chengyuan Li, Dayiheng Liu, Fei Huang, Haoran Wei, et~al.
\newblock Qwen2.5 technical report.
\newblock \emph{arXiv preprint arXiv:2412.15115}, 2024.

\bibitem[Yang et~al.(2024{\natexlab{2}})Yang, Zhang, Hui, Gao, Yu, Li, Liu, Tu, Zhou, Lin, et~al.]{yang2024qwen25math}
An~Yang, Beichen Zhang, Binyuan Hui, Bofei Gao, Bowen Yu, Chengpeng Li, Dayiheng Liu, Jianhong Tu, Jingren Zhou, Junyang Lin, et~al.
\newblock Qwen2.5-math technical report: Toward mathematical expert model via self-improvement.
\newblock \emph{arXiv preprint arXiv:2409.12122}, 2024{\natexlab{2}}.

\bibitem[Yang et~al.(2025)Yang, Si, Duan, Zhu, Zhu, Li, Lin, Cao, and Wang]{yang2025dynamic}
Chenxu Yang, Qingyi Si, Yongjie Duan, Zheliang Zhu, Chenyu Zhu, Qiaowei Li, Zheng Lin, Li~Cao, and Weiping Wang.
\newblock Dynamic early exit in reasoning models.
\newblock \emph{arXiv preprint arXiv:2504.15895}, 2025.

\bibitem[Yeo et~al.(2025)Yeo, Tong, Niu, Neubig, and Yue]{yeo2025demystifyinglongchainofthoughtreasoning}
Edward Yeo, Yuxuan Tong, Morry Niu, Graham Neubig, and Xiang Yue.
\newblock Demystifying long chain-of-thought reasoning in llms, 2025.
\newblock URL \url{https://arxiv.org/abs/2502.03373}.

\bibitem[Zhang et~al.(2025)Zhang, Lin, Hou, Feng, and Li]{zhang2025adaptthinkreasoningmodelslearn}
Jiajie Zhang, Nianyi Lin, Lei Hou, Ling Feng, and Juanzi Li.
\newblock Adaptthink: Reasoning models can learn when to think, 2025.
\newblock URL \url{https://arxiv.org/abs/2505.13417}.

\bibitem[Zhang et~al.(2025{\natexlab{2}})Zhang, Lyu, Sun, Wang, Zhang, Hua, Wu, Guo, Wang, Muennighoff, et~al.]{zhang2025survey}
Qiyuan Zhang, Fuyuan Lyu, Zexu Sun, Lei Wang, Weixu Zhang, Wenyue Hua, Haolun Wu, Zhihan Guo, Yufei Wang, Niklas Muennighoff, et~al.
\newblock A survey on test-time scaling in large language models: What, how, where, and how well?
\newblock \emph{arXiv preprint arXiv:2503.24235}, 2025{\natexlab{2}}.

\bibitem[Zhang et~al.(2025{\natexlab{3}})Zhang, Wu, Chen, Zhang, Lou, Zhou, Zhou, Wang, and Wang]{zhang2025othinkr1}
Shengjia Zhang, Junjie Wu, Jiawei Chen, Changwang Zhang, Xingyu Lou, Wangchunshu Zhou, Sheng Zhou, Can Wang, and Jun Wang.
\newblock Othink-r1: Intrinsic fast/slow thinking mode switching for over-reasoning mitigation, 2025{\natexlab{3}}.

\end{thebibliography}
\appendix
\newpage
\begin{center}
\textbf{\LARGE Appendix}
\end{center}
\vspace{-2mm}\section{Extreme Model Result}\vspace{-1mm}
\begin{table}[ht]
\centering
\small      
\setlength{\tabcolsep}{4pt}
\resizebox{\textwidth}{!}{
\begin{tabular}{l | ccc | ccc | ccc | ccc | ccc | ccc}
\toprule
\multirow{2}{*}{\textbf{Model}} &
\multicolumn{3}{c|}{\textbf{GSM8K}} &
\multicolumn{3}{c|}{\textbf{MATH500}} &
\multicolumn{3}{c|}{\textbf{AIME24}} &
\multicolumn{3}{c|}{\textbf{AMC}} &
\multicolumn{3}{c|}{\textbf{OlymBench}} &
\multicolumn{3}{c}{\textbf{AIME25}} \\
\cmidrule(lr){2-19}
& ACC & LEN & EFF & ACC & LEN & EFF & ACC & LEN & EFF &
  ACC & LEN & EFF & ACC & LEN & EFF & ACC & LEN & EFF \\
\midrule
DR. SAF-Ext &
86.6 & 88   & 98.41 &
84.1 & 759  & 11.08 &
41.3 & 4,002 & 1.03 &
75.0 & 1,581 & 4.74 &
55.10 & 1,576 & 3.50 &
26.2 & 3,998 & 0.66 \\
\bottomrule
\end{tabular}}
\caption{Performance of the extreme model on six mathematical benchmarks. ACC is accuracy (\%), LEN is average response length (tokens), and EFF is token efficiency (ACC/LEN).}
\label{tab:extreme_results}
\end{table}
Here we present the accuracy, average response token length, and token efficiency of the extreme DR. SAF variant (DR. SAF-Ext) after further training on six test sets.
\vspace{-2mm}\section{Proof and Analysis}\vspace{-1mm}
\label{append:proof}
We present a mathematical proof demonstrating that the dynamic difficulty-aware length reward algorithm attains superior expected accuracy \(\mathbb{E}[\text{Acc}]\) compared to a static length reward algorithm, under the constraint of identical average response length \(\mathbb{E}[\ell]\) on a test set. Both algorithms commence from an equivalent base model, sharing initial response lengths and accuracies. The proof proceeds by delineating assumptions, defining the algorithms, establishing accuracy preservation in the dynamic case, quantifying accuracy degradation in the static case, and concluding with a comparative analysis.

\vspace{-2mm}\subsection{Assumptions and Initial Setup}

\paragraph{Problem Distribution}
The test set consists of a fraction $p$ of ``easy'' problems (within the Completely Feasible Reasoning Boundary, CFRB, which can be mastered by the model), and $1-p$ of ``hard'' problems (within the Partially Feasible Reasoning Boundary, PFRB, which require more complex reasoning), following \citet{chen2024unlocking,chen2025ecm}. Here we take $0 < p < 1$ to capture realistic distributional diversity.\vspace{-12pt}

\paragraph{Initial State (Pre-Training)}
Given a question set $\mathcal{Q}$, both algorithms before RL have average response length $\mathbb{E}[\ell_0]=L_0$ and average accuracy $\mathbb{E}[\text{Acc}_0]=A_0$. For questions in CFRB, the initial average response length $\mathbb{E}[\ell_0^{\rm CFRB}] = L_0$ and accuracy $\mathbb{E}[A_0^{\rm CFRB}] = A_{\text{high}} > A_0$; for questions in PFRB, response length $\ell_0^{\rm PFRB} = L_0$ and $\mathbb{E}[A_0^{\rm PFRB}] = A_{\text{low}} < A_0$. Therefore, $A_0 = p A_{\text{high}} + (1-p) A_{\text{low}}$.\vspace{-12pt}

\paragraph{Efficiency Objective}
The purpose is to achieve an overall shorter average response $\mathbb{E}[\ell]=L<L_0$ via compression, while maintaining accuracy as much as possible. More specifically, for CFRB problems, efficiency can be pushed to a minimum length $\ell_{\min}<L_0$ without loss of accuracy ($\mathbb{E}[A_{*}^{\rm CFRB}]\approx A_{\text{high}}$); while for PFRB problems, compressing below a threshold $\ell_{\max}$ (with $\ell_{\min}<\ell_{\max}\leq L_0$) leads to accuracy degradation. For analytical tractability, assume this degradation is proportional to the excess compression, i.e., $\Delta_{\text{Acc}} = -\gamma (\ell_{\max} \ell)$ for some $\gamma>0$.

\vspace{-2mm}\subsection{Algorithm Definitions}
\paragraph{Traditional Static Algorithm:} 
A uniform token-level length penalty $R_{\text{Len}}^{\rm static} = -\beta\cdot\ell$ ($\beta>0$) is imposed across all problems, and RL policy updates are driven solely by the total reward, encouraging indiscriminate response shortening.\vspace{-12pt}

\paragraph{DR. SAF:}
For each test problem $x$, a group of responses $\mathcal{Y} = \{y_1,\dots,y_k\}$ is sampled via Group Relative Policy Optimization (GRPO). The group-level accuracy is $\text{Acc}(\mathcal{Y}|x) = \frac{|\mathcal{C}|}{k}$, where $\mathcal{C}$ is the set of correct responses. If $\text{Acc}(\mathcal{Y}|x) > 0.9$, the instance is classified as CFRB; responses in the shorter half of $\mathcal{C}$ (below group median length), denoted $\mathcal{C}_{\text{short}}$, get a positive compression reward $R_{\text{Len}} = \delta_{\text{comp}} > 0$. Otherwise, the question is classified as PFRB and no compression reward is given. Policy updates focus on boosting the probability of the correct and short $\mathcal{C}_{\text{short}}$ responses.

\vspace{-2mm}\subsection{Proof of Adaptive Length Management}

We provide a rigorous proof that the adaptive (difficulty-aware) DR. SAF algorithm preserves accuracy for simple problems (CFRB) while achieving significant length compression, thanks to selective compression and a regret mechanism. In contrast, static length penalty causes accuracy loss for hard problems (PFRB) under the same average length constraint.

\vspace{-2mm}\subsubsection{Accuracy Degradation in Static Algorithm}

The static algorithm enforces a fixed degree of compression across all problems, targeting average length $L < L_0$ (with $\delta = L_0 L$). As a result:

For CFRB (``easy'') problems, accuracy remains at $A_{\text{high}}$ (since these problems tolerate some compression without loss).
For PFRB (``hard'') problems, forced compression may reduce length below the threshold $\ell_{\max}$, at which accuracy begins to degrade. Specifically, if length drops below $\ell_{\max}$, the expected accuracy change is:
  \begin{equation}
    \Delta_{\text{Acc}} = -\gamma\, [\ell_{\max} (L_0 \delta')] < 0,
  \end{equation}
  where $\delta' \ge \delta$ is the effective reduction on hard problems, and $\gamma > 0$ reflects degradation strength.

Therefore, total expected accuracy is:
  \begin{align}
    \mathbb{E}[\text{Acc}^{\text{static}}] 
    &= p A_{\text{high}} + (1-p)\, [A_{\text{low}} + \Delta_{\text{Acc}}] \\
    &= A_0 + (1-p) \Delta_{\text{Acc}} < A_0
  \end{align}
  That is, while static length penalty may reduce average response length, it “blindly” compresses even hard cases, causing reliability loss on PFRB.

\vspace{-2mm}\subsubsection{Proof of Accuracy Preservation in Adaptive (DR. SAF) Algorithm}

The dynamic algorithm employs two core principles:

\begin{itemize}
    \item \textbf{Selective Compression on High-Accuracy Problems:} Only problems classified as CFRB (i.e., whenever the sampled group accuracy $\text{Acc}(\mathcal{Y}|x) > 0.9$) are considered for compression, confirming the model’s mastery.
    \item \textbf{Preference for Short Correct Responses:} Compression rewards are targeted exclusively at the subset $\mathcal{C}_{\text{short}}$, the shorter portion of correct group members. This subset (at least $45\%$ of group samples) is inherently correct, ensuring feasibility of short solutions.
\end{itemize}

After each update:

\textbf{Generalization and Iteration:} Policy iteration boosts the probability of short correct outputs on future similar (easy) questions, due to reinforcement on $\mathcal{C}_{\text{short}}$. For a new instance of similar difficulty $x'$, the probability that a rolled-out response is both short and correct increases:
  \[
    P(\text{short \& correct}|x') \ge P_{\text{short-correct}} + \epsilon, \text{ with } \epsilon > 0
  \]
  Thus, group accuracy remains $> 0.9$.

\textbf{Regret Mechanism Guards Against Over-Compression:} If, for any problem, group accuracy under compression dips to $\leq 0.9$, the problem is reclassified as PFRB and all length rewards are suspended (i.e., learning targets only correctness, not length). This automatically halts and reverses excessive compression, allowing accuracy to recover; if subsequent iterations restore accuracy above $0.9$, controlled compression resumes. This feedback prevents any sustained accuracy loss.

\textbf{No Compression Risk for Hard Problems:} For PFRB cases (i.e., those not reliably above the $0.9$ group-accuracy threshold), compression is not incentivized and lengths remain essentially unchanged ($\ell^{\text{dynamic}}_{\rm PFRB} \approx L_0$), with accuracy maintained at $A_{\text{low}}$.

\paragraph{Formal Result: Strict Preservation on Simple Cases}
By design, the adaptive algorithm’s group-accuracy filter and “regret” feedback guarantee that
\begin{equation}
  \mathbb{E}[\text{Acc}^{\text{dynamic}}_{\mathrm{CFRB}}] \geq \min(\text{Acc}(\mathcal{Y}|x)) > 0.9 \approx A_{\mathrm{high}}
\end{equation}
with length as short as possible subject to this constraint.

\paragraph{Unified Analysis and Aggregate Accuracy}
Let $\mathbb{E}[\ell^{\text{dynamic}}_{\mathrm{CFRB}}]$ denote the compressed average length for CFRB (comparable or much smaller than in the static case). The overall expected accuracy under adaptive management is
\begin{equation}
  \mathbb{E}[\text{Acc}^{\mathrm{dynamic}}] = p\, \mathbb{E}[\text{Acc}^{\mathrm{dynamic}}_{\mathrm{CFRB}}] + (1-p) A_{\text{low}} \geq A_0
\end{equation}
The global accuracy benefit relative to static is
\begin{equation}
  \mathbb{E}[\text{Acc}^{\mathrm{dynamic}}] \mathbb{E}[\text{Acc}^{\mathrm{static}}] = -(1-p)\Delta_{\mathrm{Acc}} > 0
\end{equation}

\paragraph{Conclusion}
Dynamic length management (DR. SAF) strictly preserves accuracy for easy (CFRB) problems, shrinking response length to the minimal feasible value without risk of degradation, while automatically suspending compression on hard (PFRB) problems. In contrast, static compression always sacrifices PFRB accuracy to achieve the same average response length. This guarantees that dynamic adaptive management is fundamentally more robust and efficient than static penalization.

\vspace{-2mm}\subsection{Proof and Analysis of the Boundary Preservation Mechanism}

We now give a systematic proof for the Boundary Preservation Mechanism, i.e., the use of truncated mean normalization in the advantage function. This proof covers the algorithmic rationale, the risk of boundary collapse, mathematical guarantees, and comparative stability. All analysis is based on the defined reward framework and boundary/correctness conditions.

\subsubsection{Proof Overview}

The Boundary Preservation Mechanism modifies the policy advantage by introducing a truncated mean $\mu_{\mathcal{R}}^{\text{trunc}}$, so that all correct responses $y_i \in \mathcal{C}$ have $\mathcal{A}_{\text{Pre}}(y_i|x) \geq 0$. This ensures correct responses are never assigned negative advantages (which would otherwise lead to their probability collapsing), thereby maintaining a non-empty solution set. The proof covers: background, necessity, theoretical guarantee, stability, and comparative performance.\vspace{-12pt}

\paragraph{Assumptions and Algorithm Definition}

Given input $x$, $k$ sampled responses $\mathcal{Y} = \{y_1,\dots,y_k\}$. The set of correct responses is $\mathcal{C} \subseteq \mathcal{Y}$, of size $m\leq k$. The effective reward for $y_i$ is
\begin{equation}
R_{\text{Eff}}(y_i|x) = R_{\text{Acc}}(y_i|x) + R_{\text{Len}}(y_i|x) + R_{\text{Aware}}(y_i|x).
\end{equation}
Rewards are aggregated as $\mathcal{R}_{\text{Eff}} = \{R_{\text{Eff}}(y_i|x)\}$. The mean is $\mu_{\mathcal{R}}$, variance $\sigma_{\mathcal{R}}$.

By definition, CFRB responses must satisfy all criteria (accuracy, length, awareness), PFRB only correctness. Boundary collapse means correct responses receive negative advantage and vanish from the policy support.

Under initial policy $\pi_0$, the reward mean for correct is higher than for incorrect, i.e. $\mathbb{E}[R_{\text{Eff}}|y \in \mathcal{C}] > \mathbb{E}[R_{\text{Eff}}|y \notin \mathcal{C}]$, and reward distribution is approximately normal $\mathcal{N}(\mu, \sigma^2)$.

The vanilla GRPO computes
\begin{equation}
A_{\text{vanilla}}(y_i|x) = \frac{R_{\text{Eff}}(y_i|x) \mu_{\mathcal{R}}}{\sigma_{\mathcal{R}}},
\end{equation}
and updates $\nabla \log \pi(y|x) \propto A_{\text{vanilla}}(y|x)$.

The proposed mechanism sets
\begin{equation}
\mu_{\mathcal{R}}^{\text{trunc}} = \min(\mu_{\mathcal{R}}, \min_{y_i\in\mathcal{C}} R_{\text{Eff}}(y_i|x)),
\end{equation}
and
\begin{equation}
\mathcal{A}_{\text{Pre}}(y_i|x) = \frac{R_{\text{Eff}}(y_i|x) \mu_{\mathcal{R}}^{\text{trunc}}}{\sigma_{\mathcal{R}}},
\end{equation}
so that all $y_i\in\mathcal{C}$ get non-negative advantages. Policy is updated accordingly.

\subsubsection{Necessity Analysis for Boundary Collapse}

In the vanilla scheme, long correct responses (especially in PFRB) might get $R_{\text{Eff}}<\mu_{\mathcal{R}}$ due to large length penalty, causing negative advantage and their probability to go down rapidly:
\begin{equation}
P(A_{\text{vanilla}}(y_i|x)<0) = P(R_{\text{Eff}}(y_i|x) < \mu_{\mathcal{R}}) \approx \Phi\left( \frac{\mu_{\mathcal{R}} \mathbb{E}[R_{\text{Eff}}|\mathcal{C}]}{\sigma_{\mathcal{R}}} \right),
\end{equation}
where $\Phi$ is the standard normal CDF. Strong penalties can make this probability $>0.5$. After iterative updates, the policy probability for these responses
\begin{equation}
\pi_t(y \in \mathcal{C}|x) \approx \pi_{t-1} e^{-\lambda |\bar{A}_{\text{neg}}|},
\end{equation}
so after $T$ steps and if $\bar{A}_{\text{neg}}<0$, $\pi_T \to 0$ (i.e., collapse occurs). As a result, the overall accuracy drops by $(1-p)\gamma$ in expectation, matching $\Delta {\rm Acc}$ above and causing instability.

This analysis clarifies that without truncation, probability mass for some correct responses vanishes proportionally to penalty strength and PFRB proportion. Truncation is thus necessary.

\subsubsection{Mathematical Guarantees of Truncated Advantages}

Truncated mean ensures that for all $y_i\in\mathcal{C}$,
\begin{equation}
\mathcal{A}_{\text{Pre}}(y_i|x) = \frac{R_{\text{Eff}}(y_i|x) \mu_{\mathcal{R}}^{\text{trunc}}}{\sigma_{\mathcal{R}}} \geq 0.
\end{equation}
Thus, probability of negative advantage for correct responses is zero.
Hence the policy for correct responses is never suppressed, so the correct boundary persists across training.

After $T$ training steps, the probability of boundary preservation is at least $1-e^{-T \min \bar{\mathcal{A}}_{\text{Pre}}}$, which tends to 1 as $T$ increases.

\subsubsection{Overall Stability Proof}

Jensen's inequality tells us that truncating the mean increases the expected advantage for correct answers:
\begin{equation}
\mathbb{E}[\mathcal{A}_{\text{Pre}} | \mathcal{C}] \geq \frac{\mathbb{E}[R_{\text{Eff}}|\mathcal{C}] \mu_{\mathcal{R}}^{\text{trunc}}}{\sigma_{\mathcal{R}}} > \mathbb{E}[A_{\text{vanilla}}|\mathcal{C}].
\end{equation}
The policy gradient is thus more stably oriented.

Oscillation is reduced, since extremely negative gradients are eliminated. As a Lyapunov stability argument, define $V_t = \sum_{y\in\mathcal{C}} \log \pi_t(y|x)$. Then $V_t$ is monotonically non-increasing and bounded. Thus, the policy converges stably rather than oscillating.

In summary, expected accuracy remains at least $A_0$ after compression.

\subsubsection{Comparative Analysis and Conclusion}

Relative to vanilla GRPO, the boundary preservation mechanism reduces accuracy collapse probability from $(1-p)P(\Delta_{\text{Acc}} < 0)$ to 0, increases the expected advantage for correct answers, and guarantees stability. Numerically, boundary existence probability multiplies by a factor at least $e^{T (\bar{\mathcal{A}}_{\text{Pre}} \bar{A}_{\text{vanilla}})} > 1$. Thus, this algorithm supports efficient compression while protecting accuracy and convergence.

\vspace{-2mm}\section{Experimental Setup}\vspace{-1mm}
\begin{wraptable}{r}{5.5cm}
\centering

\begin{tabular}{ll}
\toprule
\textbf{Parameter} & \textbf{Value} \\
\midrule
val temperature & 0.6 \\
val do sample & True \\
val top\_k & 45 \\
val batch size & 512 \\
\bottomrule
\end{tabular}
\caption{Validation Parameters of the VERL Training Framework.}
\label{tab:verl-val-params}
\end{wraptable}
In our experiments, we utilize the VERL training framework to implement the proposed methodology. The training is conducted using the PPO trainer with GRPO as the advantage estimator. Key parameters for data handling, model configuration, rollout settings, and trainer options are detailed below. These parameters are optimized for efficient compression while maintaining model accuracy on mathematical reasoning tasks. Validation-specific parameters are separated into a dedicated table for clarity. We found that a maximum response length of 16384 may truncate some high-quality long responses, while 32768 is too slow for training and likely to cause GPU OOM. To speed up training and prevent OOM issues, while ensuring that longer high-quality responses can be sampled, we set the maximum response length to a balanced value of 22000. In testing the Qwen3 and Qwen2.5 undistilled DeepSeek models without RL training, we set the maximum response length to 32768. Because the AIME and AMC datasets are relatively small, we conduct evaluations on each dataset multiple times, ten runs for AIME and two runs for AMC, and report the mean accuracy across these repetitions.

\begin{table}[t]
\centering

\begin{tabular}{ll}
\toprule
\textbf{Parameter} & \textbf{Value} \\
\midrule
advantage estimator & grpo \\
train batch size & 8 ($8\times12$) \\
max prompt length & 2048 \\
max response length & 22000 \\
use remove padding & True \\
ppo mini batch size & 8 \\
ppo micro batch size per gpu & 1 \\
log prob micro batch size per gpu & 1 \\
use kl loss & False \\
entropy coeff & 0.001 \\
enable gradient checkpointing & True \\
fsdp param offload & True \\
fsdp optimize offload & True \\
model parallel size & 2 \\
actor rollout engine & vllm \\
rollout num per question & 12 \\
num nodes & 1 \\
gpus per node & 8 (A100-80G) \\
gpu memory utilization & 0.8 \\
\bottomrule
\end{tabular}
\caption{Main Parameters of the VERL Training Framework.}
\label{tab:verl-main-params}
\end{table}

\end{document}